\documentclass{article}

\PassOptionsToPackage{numbers, compress}{natbib}

\usepackage[main, final]{neurips_2026}
\usepackage[utf8]{inputenc} 
\usepackage[T1]{fontenc}    
\usepackage{hyperref}       
\usepackage{url}            
\usepackage{booktabs}       
\usepackage{amsfonts}       
\usepackage{nicefrac}       
\usepackage{microtype}      
\usepackage{graphicx}       
\usepackage{subcaption}     
\usepackage{multirow}       
\usepackage{amsmath}        

\renewcommand{\arraystretch}{0.95}

\usepackage[dvipsnames,table]{xcolor}    
\definecolor{bestcolor}{RGB}{207,234,215}    
\definecolor{secondcolor}{RGB}{255,242,204}  
\definecolor{thirdcolor}{RGB}{255,224,179}   
\newcommand{\best}[1]{\cellcolor{bestcolor}\textbf{#1}}
\newcommand{\second}[1]{\cellcolor{secondcolor}#1}

\definecolor{draftcolor}{RGB}{0,0,0}

\title{StableHand: Quality-Aware Flow Matching for World-Space Dual-Hand Motion Estimation from Egocentric Video}

\author{%
  Huajian Zeng$^{1}$
  \quad
  Chaohua Yao$^{2}$
  \quad
  Yuantai Zhang$^{1}$
  \quad
  Jiaqi Yang$^{1,3}$ \\[1ex]
  \bfseries
  Rolandos Alexandros Potamias$^{4}$
  \quad
  Xingxing Zuo$^{1,*}$
  \\[2ex]
  \normalfont
  $^{1~}$Mohamed bin Zayed University of Artificial Intelligence \\
  $^{2~}$University of Illinois at Urbana-Champaign
  \quad
  $^{3~}$ShanghaiTech University \\
  $^{4~}$Imperial College London
}

\newif\ifrevmark
\revmarkfalse
\newcommand{\rev}[1]{\ifrevmark{\color{red}#1}\else#1\fi}
\newcommand{\revcolor}{\ifrevmark\color{red}\fi}
\usepackage[normalem]{ulem}
\newcommand{\del}[1]{\ifrevmark{\color{red}\sout{#1}}\fi}
\newcommand{\revb}[1]{\ifrevmark{\color{blue}#1}\else#1\fi}
\newcommand{\revbcolor}{\ifrevmark\color{blue}\fi}
\newcommand{\delb}[1]{\ifrevmark{\color{blue}\sout{#1}}\fi}

\hypersetup{hidelinks}

\begin{document}

\maketitle

\renewcommand{\thefootnote}{\fnsymbol{footnote}}
\footnotetext{Project page: \url{https://huajian-zeng.github.io/projects/stablehand}}
\footnotetext[1]{Corresponding author. (Email: {\tt\small xingxing.zuo@mbzuai.ac.ae})}
\renewcommand{\thefootnote}{\arabic{footnote}}
\setcounter{footnote}{0}

\begin{figure}[h]
  \centering
  \includegraphics[width=\linewidth]{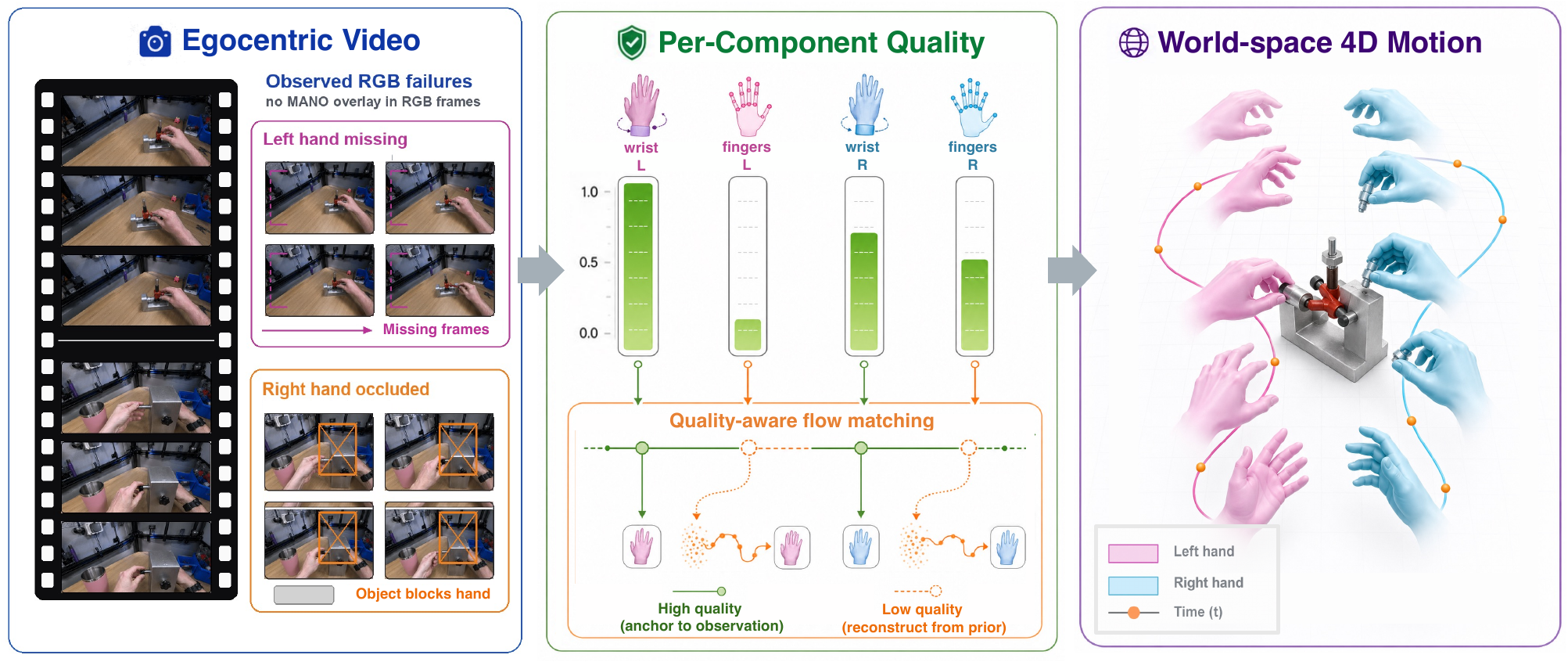}
  \caption{\textbf{StableHand recovers world space dual-hand motion from egocentric video.}
  We introduce StableHand, a quality-aware flow-matching framework driven by a per-component (wrist and fingers), per-hand quality signal $\mathbf{q}\in[0,1]^{4}$ (middle).
  Given egocentric inputs with missing or occluded hands (left), StableHand anchors reliable hand observations from a hand pose estimator and regenerates unreliable ones from a learned bimanual motion prior, yielding consistent world space dual-hand trajectories (right).}
  \label{fig:teaser}
\end{figure}

\begin{abstract}
Recovering \rev{world space} 4D motion of two interacting hands from egocentric video \del{is a fundamental capability for supervising}\rev{provides a key supervision component for} robot policy learning, where wrist trajectories track the end-effector and finger articulations specify the grasp pose.
Two major challenges arise in this setting: hands frequently leave the camera view for extended periods due to head motion, and persistent \rev{hand-object} interactions cause severe occlusions of one or both hands.
Existing methods uniformly condition on noisy hand motion observations without accounting for their per-frame reliability, leading to substantial performance degradation. Our key insight is that accurate \rev{world-space} hand motion \rev{estimation} is tightly coupled with the quality of per-frame hand observations. To this end, we decompose the quality of hand motion observations extracted from an off-the-shelf hand pose estimator into four channels: wrist global translation and finger articulations for both \rev{hands}.
We propose StableHand, a quality-aware flow-matching framework conditioned on these four-channel quality signals, which are predicted by a learned quality network. We naturally incorporate the quality signals into the flow-matching process through a per-channel forward schedule, a quality-adjusted velocity target, AdaLN modulation of the DiT denoiser, and a quality-aware ODE initialization.
This unified generative process preserves high-quality observations while reconstructing unreliable ones using a learned bimanual motion prior.
Experiments on HOT3D and ARCTIC, two egocentric benchmarks featuring long missing-hand spans and persistent \rev{hand-object} occlusions, show that StableHand achieves state-of-the-art performance across all \del{reported metrics}\rev{metrics of the main comparisons}, reducing W-MPJPE by $20$--$25\%$ compared to the strongest baseline, with the largest gains on heavily occluded ARCTIC sequences.
\end{abstract}

\section{Introduction}
\label{sec:intro}

Robot policy learning from large-scale egocentric video has scaled to dexterous and bimanual manipulation~\cite{li2025maniptrans,zheng2026egoscale}, mobile manipulation~\cite{zhu2026emma}, and vision-language-action pretraining~\cite{yang2025egovla,hoque2025egodex}, with accurate world space 4D motion of two interacting hands as a key supervision channel~\cite{li2025scalable,kareer2025egomimic,guzey2025dexterity}.
On the robot end, the wrist trajectory drives the end-effector while the finger articulation specifies the contact and grip pose required for dexterous interaction.
\rev{We target this signal as a supervision component that is used after the hand, object, scene, and robot estimates are registered to a common frame, rather than as a trajectory that composes directly with independently estimated quantities.}
%
The egocentric capture setting poses significant challenges for recovering this signal: coupled head and hand motion often causes one or both hands to exit the camera frustum for prolonged intervals, while sustained \rev{hand-object} interactions during dexterous bimanual manipulation introduce asymmetric occlusions that degrade hand observations.

%
For recovering accurate \rev{dual-hand} motions in world space, 
SLAM-based pipelines~\cite{pavlakos2024reconstructing,potamias2025wilor,valassakis2024handdgp,prakash20243d} directly transform per-frame camera-frame hand poses into the world frame via monocular SLAM~\cite{teed2021droid} without further refinement, drifting catastrophically across long missing-hand spans for lack of any temporal prior.
Existing native world space methods~\cite{zhang2025hawor,yu2025dyn,sun2026unihand} learn a sequence-level motion prior that jointly reasons about hand and camera, but condition uniformly on visual features and ignore the fact that wrist global drift is several times larger than finger articulation error, and that bimanual occlusion degrades the two hands asymmetrically.

In this work, we propose \textbf{StableHand} (Fig.~\ref{fig:teaser}), a quality-aware flow-matching framework that synthesizes the world space 4D motion of two interacting hands from egocentric video.
The framework is conditioned on a four-channel quality signal, predicted at inference by a learned quality network from hand observations extracted from an off-the-shelf hand-pose estimator.
%
%
Our key insight is that observation quality heterogeneity should serve as an explicit conditioning signal of the generative process: a single generative process can then anchor high-quality channels on the observation while regenerating low-quality channels from the prior.

Realizing this insight raises three central challenges:
\textbf{(i)~Heterogeneous quality decomposition.}
A single scalar quality per hand would average wrist global drift with finger articulation error and \rev{ignore} asymmetric bimanual occlusion, informing our four-channel signal indexed by hand and component.
\textbf{(ii)~Per-channel flow-matching pathway.}
Truncated reverse processes~\cite{meng2021sdedit,lugmayr2022repaint} and learned multivariate noise schedules~\cite{sahoo2024diffusion} fail to anchor high-quality channels while regenerating low-quality ones, motivating our per-channel forward schedule whose induced velocity target vanishes on frozen channels and a quality-aware initialization that mirrors the training-time forward.
\textbf{(iii)~Quality signal prediction at inference.}
Since the quality signal is derived from ground-truth joints at training time and unavailable at deployment, we pretrain a quality network on egocentric bimanual hand-pose corpora to predict it from observed hand parameters, hand-confidence flags, and camera pose.

In summary, our contributions are:
\begin{itemize}
  \setlength{\itemsep}{1pt}
  \setlength{\topsep}{2pt}
  \item We frame egocentric world space dual-hand recovery as per-component, quality-aware generation, identifying the two-axis (component, hand) heterogeneity of observation quality as the structure prior pipelines leave implicit.
  \item We design a quality-aware flow-matching pathway with four coupled mechanisms driven by the same four-value quality signal: a per-channel forward schedule, a quality-adjusted velocity target, AdaLN modulation of a Diffusion Transformer (DiT), and a quality-aware ODE initialization.
  \item We introduce a calibrated four-channel quality signal with a corpus-adaptive radial-basis bandwidth, predicted at inference from the egocentric visual stream by a learned quality network.
  \item On HOT3D and ARCTIC, two egocentric benchmarks targeting long missing-hand spans and persistent hand-object occlusion, StableHand achieves state-of-the-art world space dual-hand motion estimation across  \del{reported metric}\rev{most reported metrics}, with $20$--$25\%$ W-MPJPE reductions over the strongest baselines and the largest margins on the most occlusion-affected clips.
\end{itemize}

\section{Related Work}
\label{sec:related}

\noindent\textbf{Egocentric Hand Recovery.}
MANO~\cite{MANO:SIGGRAPHASIA:2017} introduced a low-dimensional parametric hand model that enabled the single-image hand regressor~\cite{boukhayma20193d} and inspired direct vertex prediction~\cite{lin2021end,lin2021mesh}.
Recent transformer-based pipelines such as HaMeR~\cite{pavlakos2024reconstructing} and WiLoR~\cite{potamias2025wilor} scale this paradigm with ViT backbones on internet-scale data, with specialized variants targeting egocentric viewpoints~\cite{prakash20243d}, differentiable global positioning~\cite{valassakis2024handdgp}, occlusion robustness~\cite{park2022handoccnet,fu2023deformer}, or hand-object interaction~\cite{liu2021semi,cao2021reconstructing}.
\rev{Concurrent work HaPTIC~\cite{ye2026predicting} extends an image-based hand regressor with temporal attention to predict coherent 4D hand trajectories from monocular video.}
World-space hand recovery extends this line by coupling SLAM~\cite{teed2021droid} or multi-stage optimization with a learned hand motion estimator~\cite{zhang2025hawor,yu2025dyn,sun2026unihand,duran2024hmp,fu2026egograsp,ye2026whole}, but none exposes per-component observation quality as an explicit conditioning signal of the prediction process.
Our work fills this gap with a calibrated per-component quality signal that decouples wrist global drift from finger articulation error and the two hands under asymmetric occlusion.

\noindent\textbf{Quality-Conditioned Diffusion.}
Diffusion models~\cite{ho2020denoising} and conditional flow matching~\cite{lipman2022flow}, often instantiated as DiTs~\cite{peebles2023scalable} and applied across image and motion domains~\rev{\cite{mu2025stablemotion,saroha2026egoflow,zeng2026flowhoi}}, support diverse auxiliary conditioning signals.
For visibility-aware generation, SDEdit~\cite{meng2021sdedit} and RePaint~\cite{lugmayr2022repaint} truncate or interleave the reverse process under a binary visibility distinction, while MuLAN~\cite{sahoo2024diffusion} learns a multivariate input-conditional schedule end-to-end without any per-channel quality signal.
\rev{In egocentric motion estimation, DSPoser~\cite{chi2024estimating} and EgoH4~\cite{hatano2025invisible} guide diffusion-based body and hand motion generation with uncertainty or visibility estimates defined at the level of joints or whole hands.}
None of these works factorizes the conditioning signal \rev{jointly} across anatomical components \del{or}\rev{and} hands.
Our framework instead naturally embeds a per-component quality signal into the flow-matching generative process.

\section{Methodology}
\label{sec:method}

\begin{figure}[t]
  \centering
  \includegraphics[width=1\linewidth]{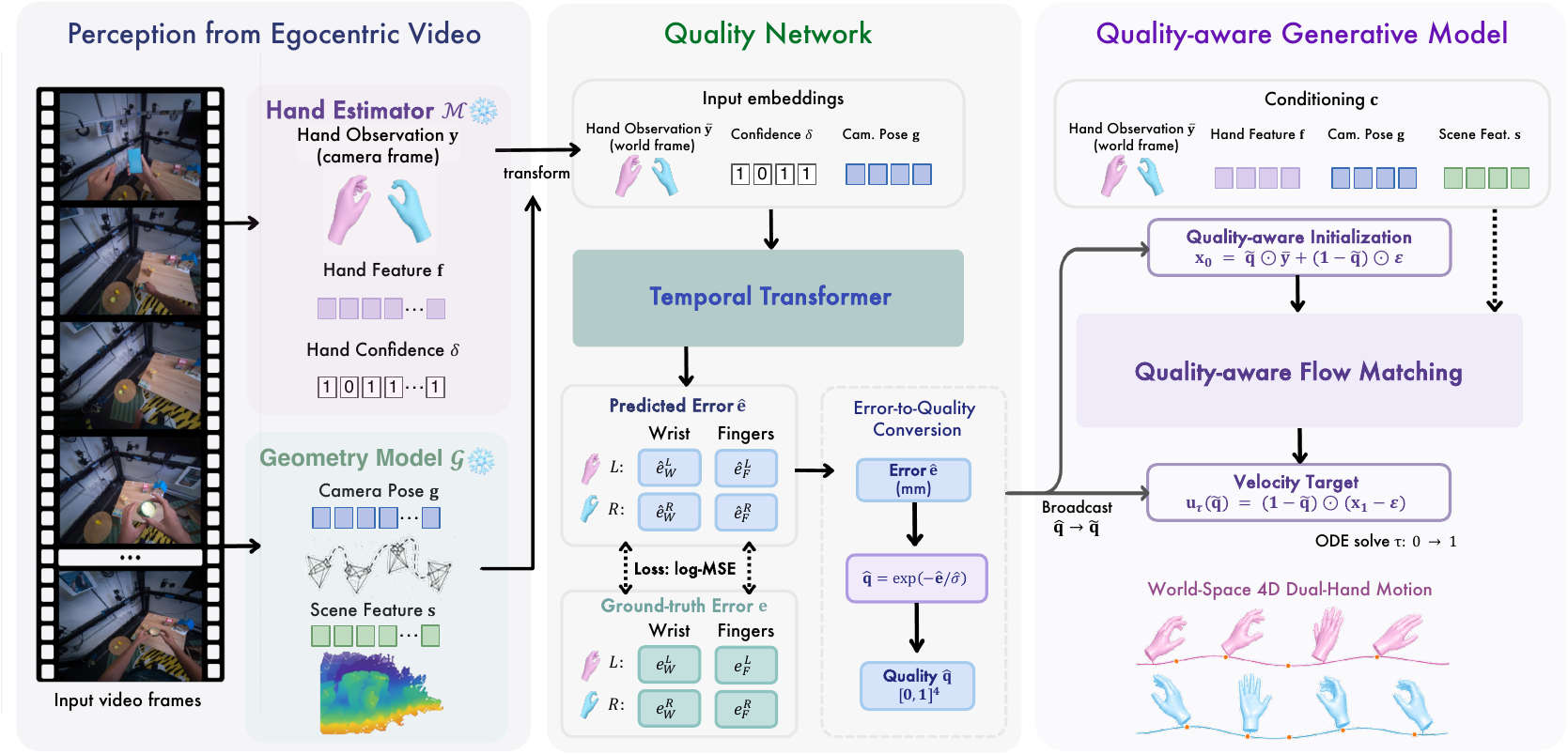}
  \caption{\textbf{StableHand pipeline.}
  From an egocentric video, frozen off-the-shelf modules~\cite{potamias2025wilor,lin2025depth} produce per-hand MANO observations together with camera and scene context.
  Our two learned modules are a \textbf{quality network} that predicts a per-component error $\hat{\mathbf{e}}$ (wrist and fingers of each hand) converted to a quality signal $\hat{\mathbf{q}}\in[0,1]^{T\times 4}$ via the Error-to-Quality Conversion block (Eq.~\ref{eq:deploy_sigma}), and a \textbf{quality-aware flow-matching model} that synthesizes the world space dual-hand motion $\mathbf{x}\in\mathbb{R}^{T\times 108}$.
  The broadcast signal $\tilde{\mathbf{q}}\in[0,1]^{T\times 108}$ drives four coupled mechanisms (bottom-right): a per-channel noise schedule, a quality-adjusted velocity target that freezes high-quality channels while regenerating low-quality ones, AdaLN modulation inside the DiT, and a quality-aware initialization at $\tau{=}0$.}
  %
  \label{fig:pipeline}
\end{figure}

Given an egocentric video $\mathbf{V}=\{\mathbf{I}_t\}_{t=1}^{T}$ of $T$ frames containing two possibly interacting hands under a moving first-person camera, we aim to recover the world space 4D motion of both hands.
As illustrated in Fig.~\ref{fig:pipeline}, our framework consists of two learned modules.
A quality network (Sec.~\ref{sec:method_qnet}) estimates a per-component (wrist and fingers of both hands) quality signal from the egocentric visual stream, and a quality-aware generative model (Sec.~\ref{sec:method_fm}) synthesizes the world space dual-hand trajectory conditioned on this signal.
In contrast to prior egocentric hand recovery that either trusts per-frame pose estimators~\cite{zhang2025hawor} or treats every channel of the hand pose observation identically~\cite{duran2024hmp}, this design conditions the generative process on a per-component quality signal, allowing high-quality components of the hand pose observation to anchor the trajectory while low-quality components are regenerated from the learned motion prior.
The two modules are trained in sequence: the quality network is pretrained on multi-dataset hand-pose corpora and frozen, after which the generative model is trained with the predicted quality signal $\hat{\mathbf{q}}$ from the frozen quality network as conditioning.

\noindent\textbf{Setup and notation.}
We represent the per-frame state of hand $h\in\{L,R\}$ at time $t$ as $\mathbf{x}^h_t = (\mathbf{p}^h_t,\,\mathbf{r}^h_t,\,\boldsymbol{\theta}^h_t) \in \mathbb{R}^{54}$, with $\mathbf{p}^h_t\in\mathbb{R}^{3}$ the root translation in a world frame, $\mathbf{r}^h_t\in\mathbb{R}^{6}$ a continuous 6D wrist rotation~\cite{zhou2019continuity}, and $\boldsymbol{\theta}^h_t\in\mathbb{R}^{45}$ the MANO~\cite{MANO:SIGGRAPHASIA:2017} finger axis-angle pose, yielding the dual-hand motion $\mathbf{x}\in\mathbb{R}^{T\times 108}$.
A frozen hand pose estimator $\mathcal{M}$~\cite{potamias2025wilor} extracts a per-hand camera frame MANO observation $\mathbf{y}^h_t\in\mathbb{R}^{54}$ and a visual feature $\mathbf{f}^h_t\in\mathbb{R}^{1280}$ from each frame, while a frozen geometry model $\mathcal{G}$~\cite{lin2025depth} produces a per-frame camera pose $\mathbf{g}_t\in\mathbb{R}^{9}$ (6D rotation and 3D translation) and a scene-geometry token $\mathbf{s}_t\in\mathbb{R}^{3072}$ from the full video.
The camera pose rigidly transforms each observation into the world frame to produce $\bar{\mathbf{y}}\in\mathbb{R}^{T\times 108}$, and is additionally consumed as cross-attention conditioning by the generative model.
Throughout, $t\in\{1,\ldots,T\}$ indexes a frame in the input video, while $\tau\in[0,1]$ indexes the flow-matching ODE time, with $\tau{=}1$ the clean target hand motion and $\tau{=}0$ the noise prior.

\subsection{Adaptive Quality Signal}
\label{sec:method_qsignal}

We associate every frame with a per-component quality signal $\mathbf{q}_t=[q^L_W,\,q^L_F,\,q^R_W,\,q^R_F]^\top\in[0,1]^{4}$ rating the quality of the wrist and fingers of each hand, kept as separate channels because their error scales decouple: wrist error is dominated by global translation drift on the tens-of-\rev{millimeter} scale, whereas finger error lives at the \rev{millimeter} scale of MANO articulation.
Let $\bar{\mathbf{j}}^{h}_{k},\mathbf{j}^{h}_{k}\in\mathbb{R}^{3}$ denote the observed and ground-truth 3D joint positions of hand $h$ ($k=0$ wrist, $k=1,\ldots,15$ fingers), and $\bar{\mathbf{R}}^h,\mathbf{R}^h\in SO(3)$ the corresponding wrist rotation matrices recovered from the 6D representation $\mathbf{r}^h_t$.
The wrist channel sums the wrist translation error and the wrist rotation error scaled by the canonical MANO palm radius $r_0\!=\!85$\,mm.
The finger channel uses MPJPE~\cite{ionescu2013human3} in the ground-truth wrist frame, which isolates finger articulation from wrist-orientation error via the alignment rotation $\mathbf{R}^h(\bar{\mathbf{R}}^h)^{\!\top}$:
\begin{equation}
e^h_W = \|\bar{\mathbf{j}}^{h}_{0} - \mathbf{j}^{h}_{0}\|_2 + r_0\,\arccos\!\left[\tfrac{1}{2}\bigl(\mathrm{tr}(\bar{\mathbf{R}}^{h\top}\mathbf{R}^h) - 1\bigr)\right],
\quad q^h_W = \exp(-e^h_W/\sigma_W),
\label{eq:q_wrist}
\end{equation}
\begin{equation}
e^h_F = \tfrac{1}{15}\sum_{k=1}^{15}\|\mathbf{R}^h(\bar{\mathbf{R}}^h)^{\!\top}(\bar{\mathbf{j}}^{h}_{k} - \bar{\mathbf{j}}^{h}_{0}) - (\mathbf{j}^{h}_{k}-\mathbf{j}^{h}_{0})\|_2,
\quad q^h_F = \exp(-e^h_F/\sigma_F).
\label{eq:q_finger}
\end{equation}
Each entry follows an RBF kernel~\cite{scholkopf2002learning} $q=\exp(-e/\sigma)$ over a component-specific joint error. We calibrate the RBF bandwidth per corpus so that the worst $20\%$ of frames (those with the largest errors) fall into the low-quality region $q<0.1$. Equivalently, $\sigma$ is chosen to satisfy $\exp(-e/\sigma)=0.1$ at the $80$-th-percentile error,
\begin{equation}
\sigma_\star = p_{80}(e_\star)/\ln 10, \quad \hat{\sigma}_\star = p_{80}(\hat{e}_\star)/\ln 10, \qquad \star\in\{W,F\},
\label{eq:deploy_sigma}
\end{equation}
where $\hat{\sigma}$ is recovered at deployment from the quality network's predictions $\hat{\mathbf{e}}$ on a local-data calibration subset.
Frames without a valid detection or annotation receive $q=0$.
%
%
The perturbation pool and the $20\%$-target sensitivity are in the supplemental material.

\subsection{Quality Network}
\label{sec:method_qnet}

The ground-truth $\mathbf{q}$ defined by Eqs.~\ref{eq:q_wrist}--\ref{eq:q_finger} requires joint annotations unavailable at inference, motivating a learned predictor.
In the quality network, we predict the per-component error $\hat{\mathbf{e}}$ (mm) directly rather than the quality $\hat{\mathbf{q}}$, since errors live on a physical scale common to all corpora while $\hat{\mathbf{q}}$ depends on the corpus-specific bandwidth $\sigma$.
A single $(\bar{\mathbf{y}},\delta,\mathbf{g})\!\to\!\hat{\mathbf{e}}$ mapping can therefore be learned across corpora whose $\sigma$ values span more than an order of magnitude.
At deployment, the quality signal is recovered by the Error-to-Quality Conversion (Fig.~\ref{fig:pipeline}),
\begin{equation}
\hat{\mathbf{q}}=\exp(-\hat{\mathbf{e}}/\hat{\sigma}),
\label{eq:e_to_q}
\end{equation}
with the bandwidth $\hat{\sigma}$ adaptively calibrated from QN predictions on a local data subset as in Eq.~\ref{eq:deploy_sigma}.
Hand observation quality depends on both spatial cues within a single frame (e.g., occlusion or motion blur) and temporal cues across consecutive frames (e.g., sustained detection gaps or rapid camera rotation).
To capture both, our quality network jointly processes a $T$-frame window of world space MANO observations and hand-confidence flags (Fig.~\ref{fig:pipeline}, middle panel).

\noindent\textbf{Architecture.}
At each frame $t$, the per-hand input concatenates the world space MANO observation $\bar{\mathbf{y}}^h_t$, a binary hand-confidence flag $\delta^h_t$, and the per-frame camera pose $\mathbf{g}_t$ that anchors wrist translation to head ego-motion.
A learnable \textsc{[geo]} token prepended to the temporal sequence aggregates trajectory-level context through self-attention and broadcasts it back to every per-frame token, so that each per-frame quality prediction is conditioned on the clip-level quality pattern rather than on local frames alone.
A temporal self-attention encoder with RoPE~\cite{vaswani2017attention,su2024roformer} processes the augmented sequence, after which a shared per-frame two-layer MLP outputs the log-space prediction $\log(1+\hat{\mathbf{e}}^h_t)$ that exponentiates to the per-hand error $\hat{\mathbf{e}}^h_t = (\hat{e}^h_W,\hat{e}^h_F) \in \mathbb{R}_{\geq 0}^{2}$ (mm), stacked across both hands as $\hat{\mathbf{e}}_t\in\mathbb{R}_{\geq 0}^{4}$.
The frozen hand estimator's feature $\mathbf{f}^h_t$ is deliberately excluded from the input to suppress the dataset-style shortcut it injects under multi-corpus pretraining.
Undetected frames ($\delta^h_t=0$) hard-override $\hat{\mathbf{q}}^h_t=\mathbf{0}$ at the Error-to-Quality Conversion step (Fig.~\ref{fig:pipeline}).
The ablation of the visual-feature exclusion is provided in the supplemental material.

\noindent\textbf{Training.}
The quality network is pretrained on eight egocentric bimanual hand-pose corpora (${\sim}10$M frames, Sec.~\ref{subsec:exp_setup}).
A log-space mean-squared-error term supervises per-sample magnitude over frame-hand pairs $\mathcal{V}$ with valid annotations,
\begin{equation}
\mathcal{L}_{\mathrm{MSE}} \;=\; \tfrac{1}{|\mathcal{V}|}\sum_{(t,h)\in\mathcal{V}}\bigl\|\log(1+\hat{\mathbf{e}}^h_t) - \log(1+\mathbf{e}^h_t)\bigr\|_2^2,
\label{eq:qn_mse}
\end{equation}
where $\mathbf{e}^h_t$ is the ground-truth per-component error of Eqs.~\ref{eq:q_wrist}--\ref{eq:q_finger} and the $\log(1+\cdot)$ transform balances per-corpus contributions whose error scales span over an order of magnitude.
The MSE objective biases the QN toward the conditional mean and narrows the spread of $\hat{\mathbf{e}}$ relative to the ground-truth error.
As a consequence, the deployment bandwidth $\hat{\sigma}=p_{80}(\hat{\mathbf{e}})/\ln 10$ falls below the training-time $\sigma$, and the deployment-time $\hat{\mathbf{q}}$ distribution is more polarized (concentrated near $0$ or $1$) than the one the DiT was trained on.
To restore the distribution shape that the per-sample MSE compresses, we add an auxiliary 1D Wasserstein-1 distance~\cite{ambrosio2005gradient} between the empirical distributions of $\log(1+\hat{\mathbf{e}})$ and $\log(1+\mathbf{e})$ within each batch.
For scalar variables this distance admits the closed-form L1 difference between sorted empirical samples, providing a differentiable, distribution-level signal that per-sample MSE cannot supply:
\begin{equation}
\mathcal{L}_{\mathrm{W}} \;=\; \tfrac{1}{N}\sum_{i=1}^{N} \Bigl| \mathrm{sort}_i\bigl(\log(1+\hat{\mathbf{e}})\bigr) - \mathrm{sort}_i\bigl(\log(1+\mathbf{e})\bigr) \Bigr|,
\label{eq:qn_wasserstein}
\end{equation}
where $\mathrm{sort}_i(\cdot)$ extracts the $i$-th smallest entry over the $N$ valid frame-hand-component triples in the batch.
This analytical 1D Wasserstein-1 distance aligns the full empirical CDF and matches $p_{80}(\hat{\mathbf{e}})$ to $p_{80}(\mathbf{e})$ by construction, recovering the training-time $\sigma$ at deployment.
The total quality-network loss combines the two terms,
\begin{equation}
\mathcal{L}_{\mathrm{QN}} \;=\; \mathcal{L}_{\mathrm{MSE}} + \lambda_{\mathrm{W}}\,\mathcal{L}_{\mathrm{W}},
\label{eq:qn_loss}
\end{equation}
with $\lambda_{\mathrm{W}}=1.0$.
After pretraining, the quality network is frozen and its predictions $\hat{\mathbf{e}}$ are converted to $\hat{\mathbf{q}}$ via Eq.~\ref{eq:e_to_q} at both the generative model's training and inference.
The regression target is a per-component MPJPE error in physical units (Eqs.~\ref{eq:q_wrist}--\ref{eq:q_finger}), defined by dual-hand geometry and independent of the upstream estimator, so the same $(\bar{\mathbf{y}},\delta,\mathbf{g})\!\to\!\hat{\mathbf{e}}$ mapping transfers to a different regressor by retraining on its outputs and recalibrating $\hat{\sigma}$ via Eq.~\ref{eq:deploy_sigma}, which we validate by swapping WiLoR~\cite{potamias2025wilor} for HaMeR~\cite{pavlakos2024reconstructing} in the supplemental material.

\subsection{Quality-Aware Hand Motion Generation}
\label{sec:method_fm}

We instantiate the generative model as a quality-aware conditional flow-matching DiT (Fig.~\ref{fig:pipeline}, right panel) and detail its four parts below: a per-channel forward process driven by $\tilde{\mathbf{q}}_t$, the resulting training objective, a denoiser architecture with AdaLN modulation by $\tilde{\mathbf{q}}_t$, and a quality-aware inference initialization at $\tau{=}0$.

\noindent\textbf{Per-channel forward process.}
Standard conditional flow matching (CFM)~\cite{lipman2022flow} adopts a linear path $\mathbf{x}_\tau = \tau\,\mathbf{x}_1 + (1-\tau)\,\boldsymbol{\varepsilon}$ between the clean dual-hand motion $\mathbf{x}_1\in\mathbb{R}^{T\times 108}$ and noise $\boldsymbol{\varepsilon}\sim\mathcal{N}(0,I)$ with velocity target $\mathbf{u}_\tau = \mathbf{x}_1 - \boldsymbol{\varepsilon}$, whose scalar schedule treats every channel identically.

In our work, we replace this scalar schedule with a per-channel one driven by the quality signal.
We broadcast $\mathbf{q}$ onto $\tilde{\mathbf{q}}\in[0,1]^{T\times 108}$ by replicating $q^h_W$ at time $t$ onto the wrist slots $(\mathbf{p}^h_t,\mathbf{r}^h_t)$ and $q^h_F$ onto the finger slots $\boldsymbol{\theta}^h_t$ of each hand, yielding the per-channel schedule and its associated velocity target,
\begin{equation}
\alpha(\tau,\tilde{\mathbf{q}}) = 1-(1-\tau)(1-\tilde{\mathbf{q}}),
\quad
\mathbf{x}_\tau = \alpha\odot\mathbf{x}_1 + (1-\alpha)\odot\boldsymbol{\varepsilon},
\quad
\mathbf{u}_\tau(\tilde{\mathbf{q}}) = (1-\tilde{\mathbf{q}})\odot(\mathbf{x}_1 - \boldsymbol{\varepsilon})\rev{,}
\label{eq:forward}
\end{equation}
where $\odot$ denotes the element-wise (Hadamard) product.

This per-channel schedule has two informative extremes that drive the design.
For any motion channel $c$, $\tilde{q}_{t,c}=0$ recovers standard flow matching, while $\tilde{q}_{t,c}=1$ freezes channel $c$ at $\mathbf{x}_1$ via $\alpha\equiv 1$ and $\mathbf{u}\equiv 0$, supervising the denoiser to leave high-quality channels untouched while regenerating low-quality channels from noise.

\noindent\textbf{Training objective.}
A learned vector field $\mathbf{v}_\theta(\mathbf{x}_\tau,\tau;\mathbf{c})$ is trained to regress the quality-adjusted velocity target of Eq.~\ref{eq:forward} via
\begin{equation}
\mathcal{L}_{\mathrm{FM}} \;=\; \mathbb{E}_{\tau,\mathbf{x}_1,\boldsymbol{\varepsilon}}\bigl[\,\|\mathbf{v}_\theta(\mathbf{x}_\tau,\tau;\mathbf{c}) - \mathbf{u}_\tau(\tilde{\mathbf{q}})\|_2^2\,\bigr],
\label{eq:fm_loss}
\end{equation}
where $\mathbf{c}=\{\mathbf{c}_t\}_{t=1}^{T}$ collects per-frame cross-attention conditioning bundles, each $\mathbf{c}_t$ comprising the scene-geometry token $\mathbf{s}_t$~\cite{lin2025depth}, the camera pose $\mathbf{g}_t$, and the per-hand observation $\mathbf{z}^h_t=[\bar{\mathbf{y}}^h_t;\,\mathbf{f}^h_t]$ for $h\in\{L,R\}$.
On top of $\mathcal{L}_{\mathrm{FM}}$ we add a second-order temporal smoothness regularizer on the implied clean trajectory $\widetilde{\mathbf{x}}_1=\mathbf{x}_\tau+(1-\tau)\,\mathbf{v}_\theta$ (which recovers $\mathbf{x}_1$ exactly via $1{-}\alpha(\tau,\tilde{\mathbf{q}})=(1{-}\tau)(1{-}\tilde{\mathbf{q}})$) to suppress frame-to-frame jitter on channels generated from noise, and an auxiliary wrist-translation term $\mathcal{L}_{\mathrm{wrist}}$ to rebalance the 3-dimensional wrist slot otherwise dominated by the 45-dimensional finger component:
\begin{equation}
\mathcal{L}_{\mathrm{smooth}} = \tfrac{1}{T-2}\!\sum_{t=2}^{T-1}\!\|\widetilde{\mathbf{x}}_{1,t-1}{-}2\widetilde{\mathbf{x}}_{1,t}{+}\widetilde{\mathbf{x}}_{1,t+1}\|_2^2,
\quad
\mathcal{L}_{\mathrm{wrist}} = \tfrac{1}{2T}\!\sum_{t=1}^{T}\!\sum_{h\in\{L,R\}}\! \|\widetilde{\mathbf{p}}^h_{1,t} - \mathbf{p}^h_t\|_2^2,
\label{eq:aux_losses}
\end{equation}
where $\widetilde{\mathbf{p}}^h_{1,t}$ extracts the wrist translation of $\widetilde{\mathbf{x}}_{1,t}$.
The total training objective is $\mathcal{L} = \mathcal{L}_{\mathrm{FM}} + \lambda_{\mathrm{smooth}}\mathcal{L}_{\mathrm{smooth}} + \lambda_{\mathrm{wrist}}\mathcal{L}_{\mathrm{wrist}}$ with $\lambda_{\mathrm{smooth}}=0.1$ and $\lambda_{\mathrm{wrist}}=5.0$.

\noindent\textbf{Denoiser architecture.}
We instantiate $\mathbf{v}_\theta$ as a DiT~\cite{peebles2023scalable} that processes each frame as four tokens, one per quality channel of $\tilde{\mathbf{q}}_t$ (left/right wrist and left/right fingers), letting AdaLN driven by $\mathbf{m}_t = \mathrm{MLP}_\tau(\tau) + \mathrm{MLP}_q(\tilde{\mathbf{q}}_t)$ scale wrist and finger updates to match the per-channel quality structure of the forward process.
Temporal self-attention with RoPE~\cite{su2024roformer,vaswani2017attention} spans all $4T$ tokens of the $T$-frame clip, while cross-attention injects the conditioning bundle $\mathbf{c}_t$ as complementary context.

\noindent\textbf{Quality-aware inference initialization.}
At inference we replace the stochastic forward process with a \emph{deterministic mapping} from the quality-aware initialization at $\tau{=}0$ to the recovered trajectory at $\tau{=}1$, realized by integrating the deterministic reverse ODE $\mathrm{d}\mathbf{x}/\mathrm{d}\tau = \mathbf{v}_\theta(\mathbf{x},\tau;\mathbf{c})$ over $\tau\in[0,1]$ from
\begin{equation}
\mathbf{x}_{0} \;=\; \tilde{\mathbf{q}}\odot\bar{\mathbf{y}} + (1-\tilde{\mathbf{q}})\odot\boldsymbol{\varepsilon},
\qquad \boldsymbol{\varepsilon}\sim\mathcal{N}(0,I),
\label{eq:qs_init}
\end{equation}
which matches the training-time $\mathbf{x}_\tau$ of Eq.~\ref{eq:forward} at $\tau{=}0$ since $\alpha(0,\tilde{\mathbf{q}})=\tilde{\mathbf{q}}$.
We initialize each channel of the reverse ODE at the noise level matching its own quality, exactly as the training-time forward of Eq.~\ref{eq:forward} places it at $\tau{=}0$: high-quality channels start near the observation and low-quality channels near pure noise.
SDEdit-style~\cite{meng2021sdedit} truncated reverse processes instead start every channel from a single intermediate noise level, which cannot reproduce the per-channel structure of the training-time forward (each channel sits at noise level $1-\tilde{q}_c$ at $\tau{=}0$).
Our initialization eliminates this per-channel train-test mismatch by construction.

\section{Experiments}
\label{sec:experiments}

We evaluate world space 4D dual-hand motion estimation against two complementary failure modes of egocentric hand observation: extended hand out-of-view spans under dynamic head-mounted camera motion (HOT3D~\cite{banerjee2024introducing}), and dexterous bimanual manipulation with persistent hand-object occlusion and asymmetric per-component quality (ARCTIC~\cite{fan2023arctic}).
Sec.~\ref{subsec:exp_main} reports quantitative comparisons, and Sec.~\ref{subsec:exp_ablation} ablates the core design choices on HOT3D.

\subsection{Experimental Setup}
\label{subsec:exp_setup}

\noindent\textbf{Datasets.}
We evaluate on HOT3D~\cite{banerjee2024introducing} and ARCTIC~\cite{fan2023arctic}.
The generative model is trained per benchmark while the quality network (Sec.~\ref{sec:method_qnet}) is shared across both.
Splits and pretraining details are in the supplemental material.

\noindent\textbf{Evaluation metrics.}
\del{Following~\mbox{\cite{ye2023decoupling,zhang2025hawor,yu2025dyn}}, we report}\rev{We report} PA-MPJPE, W-MPJPE, WA-MPJPE, and Acceleration Error (AccEr)\rev{, adopting the metric names of~\cite{ye2023decoupling,zhang2025hawor,yu2025dyn}}.
\rev{W-MPJPE fits no alignment and is computed in a gravity-aligned world frame anchored at the first-frame camera, WA-MPJPE \delb{measures the joint error after subtracting each frame's wrist position}\revb{subtracts the wrist position of each frame from the predicted and the ground-truth joints separately, without rotation or scale alignment}, and PA-MPJPE applies a per-frame Procrustes alignment.}
On ARCTIC, we additionally report MRRPE~\cite{fan2023arctic,moon2020interhand2} for the two-hand relative spatial consistency.
All metrics are computed on every GT-valid frame with no detection-failure filtering.
Precise definitions are in the supplemental material\revb{ (Sec.~\ref{supp:metrics})}.

\noindent\textbf{Baselines.}
We compare against two families of prior work, distinguished by how camera information enters the world space estimation.
\emph{SLAM-based} methods directly transform a per-frame camera-frame hand pose into the world frame via a SLAM-estimated camera pose~\cite{teed2021droid}, with no further refinement: HaMeR~\cite{pavlakos2024reconstructing}, WiLoR~\cite{potamias2025wilor}, and HMP~\cite{duran2024hmp}.
\emph{Native world space} methods learn a world-frame hand motion model that jointly reasons about hand and camera, integrating SLAM (when used) as one component of a learned sequence-level pipeline rather than as a stand-alone projection: HaWoR~\cite{zhang2025hawor}, UniHand~\cite{sun2026unihand}, and Dyn-HaMR~\cite{yu2025dyn}.
Native world space baselines are retrained on the corresponding training split using their official codebases, while SLAM-based baselines use public checkpoints.
UniHand numbers ($\ddagger$ in Tab.~\ref{table:hot3d}) are quoted from the original paper since its code is unreleased.

\noindent\textbf{Implementation details.}
Both networks are trained on a single H100 with AdamW~\cite{loshchilov2017decoupled} optimizer.
Inference runs a $20$-step Euler ODE solve over each $T{=}150$ clip on per-frame dual-hand observations from a frozen WiLoR estimator~\cite{potamias2025wilor}.
Full hyperparameters and architectural details are in the supplemental material.

\subsection{Comparison on HOT3D and ARCTIC}
\label{subsec:exp_main}

\begin{table}[t]
\centering
\caption{\textbf{World-space hand motion estimation on HOT3D~\cite{banerjee2024introducing}.}
The upper block reports SLAM-based methods.
The lower block reports native world space methods.
$\ddagger$~numbers taken from the original paper (code not publicly available).
Per-column \colorbox{bestcolor}{\textbf{best}} and \colorbox{secondcolor}{second best} are color-coded.}
\label{table:hot3d}
\small
\setlength{\tabcolsep}{3pt}
\begin{tabular}{lcccc}
\toprule
Method
  & PA-MPJPE [mm] ($\downarrow$)
  & W-MPJPE [mm] ($\downarrow$)
  & WA-MPJPE [mm] ($\downarrow$)
  & AccEr [m/s$^2$] ($\downarrow$) \\
\midrule
HaMeR-SLAM~\cite{pavlakos2024reconstructing}          & 9.07  & 295.32 & 73.90 & 13.81 \\
WiLoR-SLAM~\cite{potamias2025wilor}                   & 8.60  & 171.24 & 54.34 & 12.30 \\
HMP-SLAM~\cite{duran2024hmp}                          & 11.72 & 131.19 & 42.77 & 5.83  \\
\midrule
Dyn-HaMR~\cite{yu2025dyn}                        & 8.64  & 86.78  & 40.47 & 5.46  \\
HaWoR~\cite{zhang2025hawor}                      & 8.65  & 71.89  & 30.20 & 7.80  \\
UniHand$^\ddagger$~\cite{sun2026unihand}           & \second{4.76}  & \second{63.97}  & \second{25.24} & \second{4.93}  \\
\textbf{Ours}                                     & \best{4.02}  & \best{57.83}  & \best{21.02} & \best{3.83} \\
\bottomrule
\end{tabular}
\end{table}

\begin{figure*}[!t]
  \centering
  \includegraphics[width=0.95\linewidth]{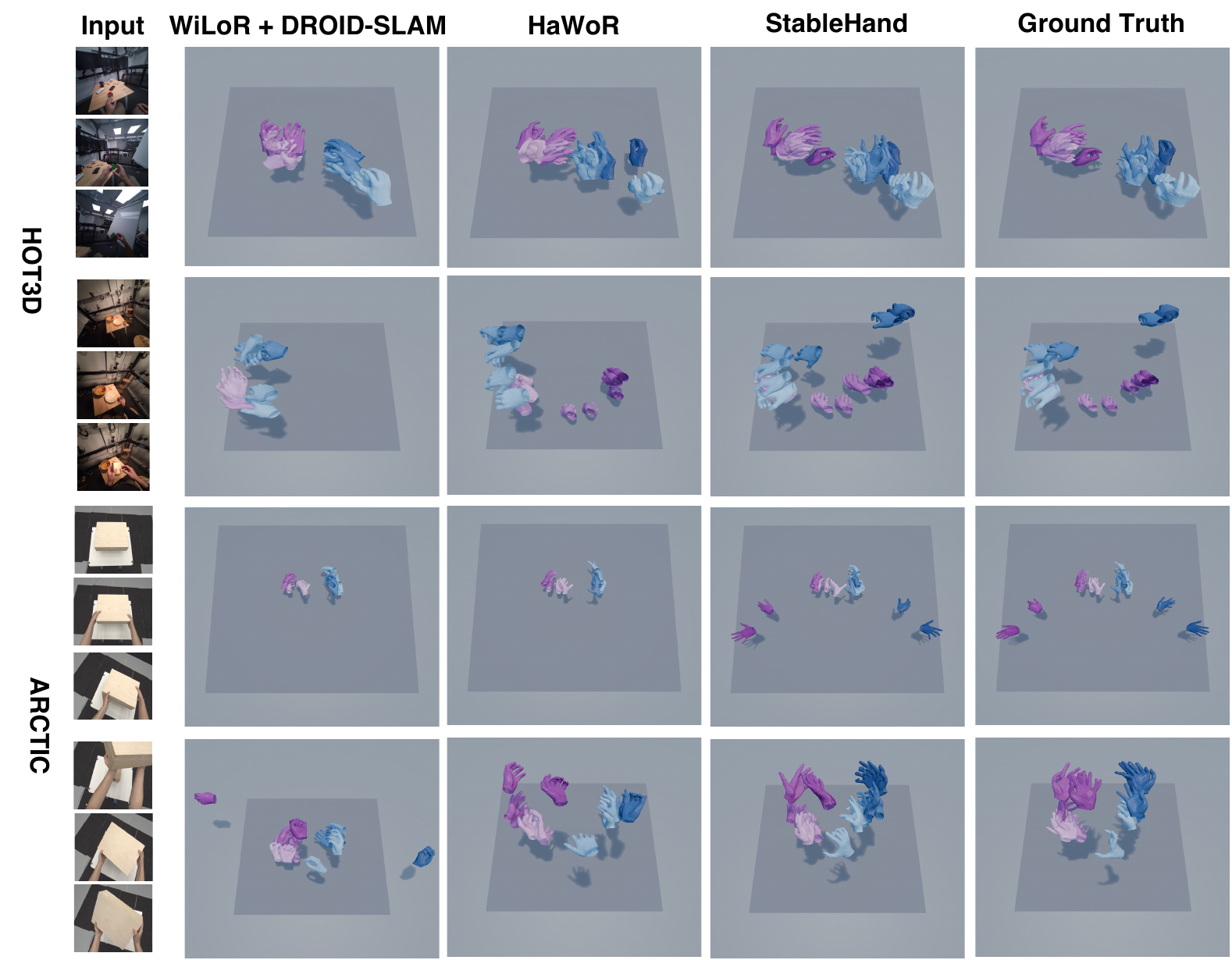}
  \caption{\textbf{Qualitative comparison on HOT3D (top two rows, long missing-hand spans) and ARCTIC (bottom two rows, persistent hand-object occlusion).}
  Each row shows three input frames and the world space dual-hand mesh trajectory of WiLoR~\cite{potamias2025wilor}+DROID-SLAM~\cite{teed2021droid}, HaWoR~\cite{zhang2025hawor}, our StableHand, and Ground Truth (left hand magenta, right hand blue, with mesh shading dark$\to$light encoding temporal order).
  StableHand preserves coherent trajectories and re-anchors when observations resume, while WiLoR+DROID-SLAM drifts and HaWoR over-smooths or collapses the occluded hand to a generic prior.
  Best viewed in the supplementary video.
  }
  \label{fig:qualitative}
\end{figure*}

\begin{figure}[!t]
  \centering
  \begin{subfigure}{0.48\linewidth}
    \centering
    \includegraphics[width=\linewidth]{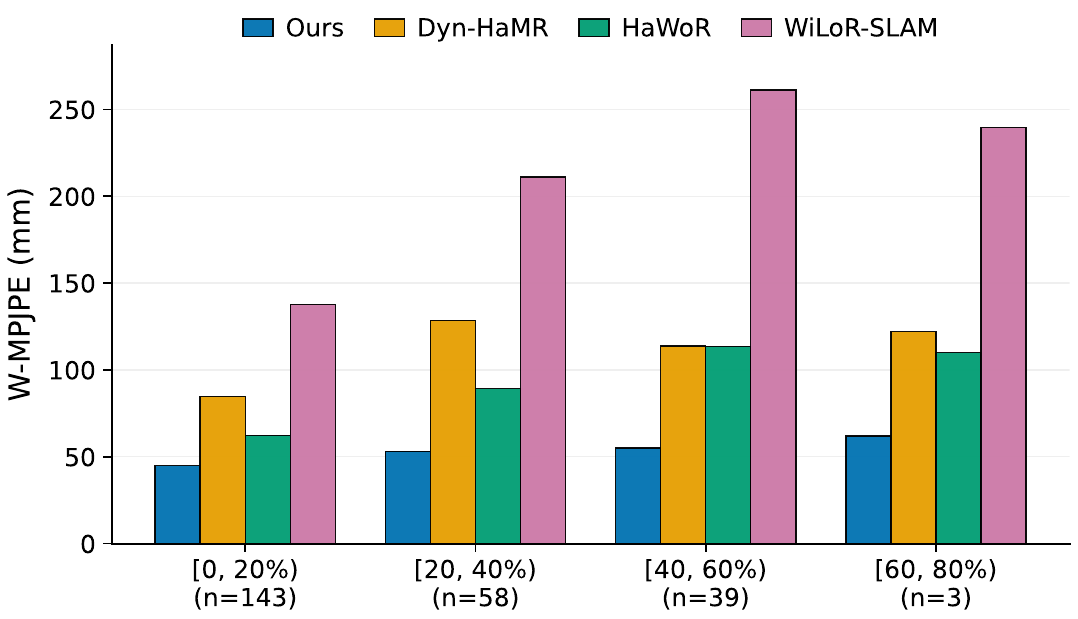}
    \caption{HOT3D, by per-clip missing-hand fraction}
    \label{fig:hot3d_stratified}
  \end{subfigure}
  \hfill
  \begin{subfigure}{0.48\linewidth}
    \centering
    \includegraphics[width=\linewidth]{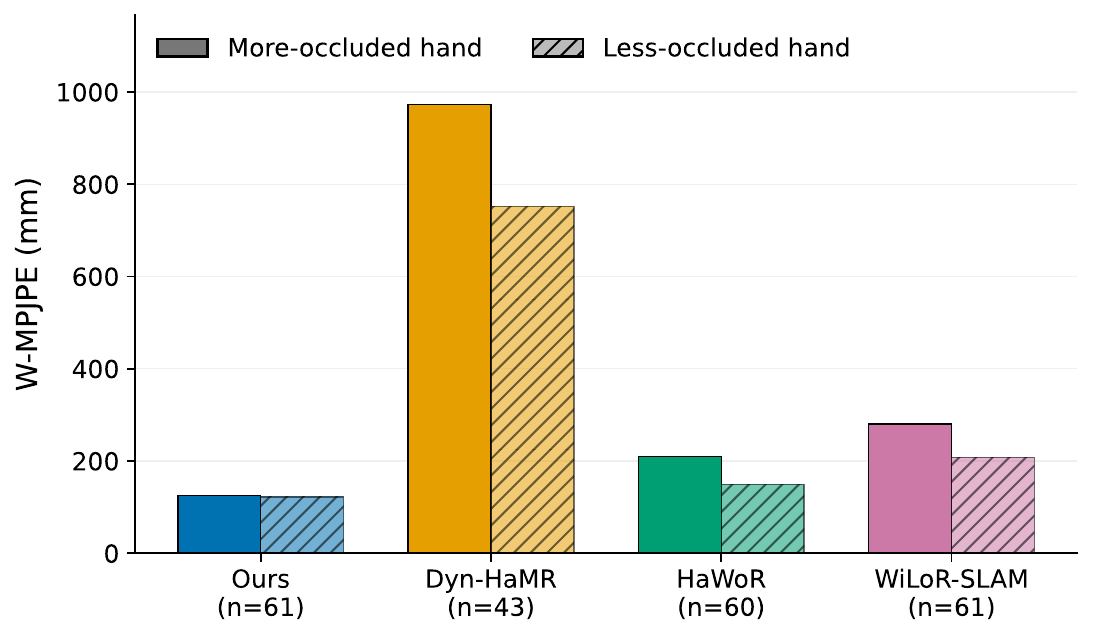}
    \caption{ARCTIC, per-hand under detector failure}
    \label{fig:arctic_asymmetry}
  \end{subfigure}
  \caption{\textbf{Stratified evaluations.} (a)~On HOT3D our W-MPJPE is lowest in every bin, and the gap to WiLoR-SLAM~\cite{potamias2025wilor}, Dyn-HaMR~\cite{yu2025dyn}, and HaWoR~\cite{zhang2025hawor} is largest in the high-missing regime. The lower baseline values in the $60$--$80\%$ bin reflect small-sample statistics (per-bin clip counts annotated above each bar). (b)~On ARCTIC, single-hand baselines exhibit a substantial W-MPJPE gap between the more- and less-occluded hand. Our model closes most of this gap, with a small residual difference reflecting that the less-occluded hand still carries cleaner observations.}
\end{figure}

\noindent\textbf{HOT3D~\cite{banerjee2024introducing}: long missing-hand spans.}
Our method achieves the lowest error on every metric of Tab.~\ref{table:hot3d}, reducing W-MPJPE by ${\sim}20\%$ over HaWoR~\cite{zhang2025hawor}, the strongest baseline with public code, while SLAM-based baselines drift catastrophically for lack of any temporal prior.
Two mechanisms drive this gap, visualized in Fig.~\ref{fig:qualitative} (top two rows): the quality network identifies missing-hand frames and the per-channel forward process regenerates them from the prior without disturbing the visible hand's observations.
Stratifying test clips by missing-hand fraction (Fig.~\ref{fig:hot3d_stratified}) shows our W-MPJPE advantage concentrates in the high-missing regime that the quality-aware schedule targets.

\begin{table}[t]
\centering
\caption{\textbf{World-space hand motion estimation on ARCTIC~\cite{fan2023arctic}.}
The upper block reports SLAM-based methods.
The lower block reports native world space methods, retrained on the ARCTIC training split.
Per-column \colorbox{bestcolor}{\textbf{best}} and \colorbox{secondcolor}{second best} are color-coded.}
\label{table:arctic}
\small
\setlength{\tabcolsep}{3pt}
\resizebox{\linewidth}{!}{%
\begin{tabular}{lccccc}
\toprule
Method
  & PA-MPJPE [mm] ($\downarrow$)
  & W-MPJPE [mm] ($\downarrow$)
  & WA-MPJPE [mm] ($\downarrow$)
  & AccEr [m/s$^2$] ($\downarrow$)
  & MRRPE [mm] ($\downarrow$) \\
\midrule
HaMeR-SLAM~\cite{pavlakos2024reconstructing}          & 10.99  & 232.48 & 59.35 & 10.82 & 163.46 \\
WiLoR-SLAM~\cite{potamias2025wilor}                   & \second{8.92}  & 183.72 & 50.03 & 8.80  & 122.42 \\
HMP-SLAM~\cite{duran2024hmp}                          & 13.73 & \second{166.21} & \second{41.62} & 7.13  & 131.33 \\
\midrule
Dyn-HaMR~\cite{yu2025dyn}                             & 13.36 & 734.31 & 83.63 & \second{6.05}  & 189.09 \\
HaWoR~\cite{zhang2025hawor}                           & 10.30 & 171.61 & 51.98 & 11.35 & \second{112.36} \\
\textbf{Ours}                                         & \best{8.07}  & \best{124.59} & \best{32.66} & \best{4.67}  & \best{93.71}  \\
\bottomrule
\end{tabular}
}
\end{table}

\noindent\textbf{ARCTIC~\cite{fan2023arctic}: persistent hand-object occlusion.}
Our method achieves the lowest error on every metric of Tab.~\ref{table:arctic}, reducing W-MPJPE by ${\sim}25\%$ over the strongest baseline, with comparable margins on WA-MPJPE and MRRPE.
The MRRPE gain reflects our dual-hand trajectory representation that lets the generative model reason jointly about the two hands' relative position.
ARCTIC reverses the HOT3D family ranking, where native world space methods suffer disproportionately because their temporal coupling propagates one hand's noisy observation into the other's trajectory (Dyn-HaMR's W-MPJPE alone increases $8.5\times$ from HOT3D to ARCTIC).
The aggregate metrics also mask a per-hand asymmetry (Fig.~\ref{fig:arctic_asymmetry}): single-hand baselines collapse the low-quality hand to a generic prior, while our model closes the per-hand gap within measurement noise via shared self-attention over all $4T$ tokens.
Fig.~\ref{fig:qualitative} (bottom two rows) visualizes this across long contact phases. Additional qualitative comparisons and failure cases for both benchmarks are provided in the supplemental material.

\noindent\textbf{\rev{Calibration-only transfer.}}
\rev{We freeze a quality network pretrained without H2O and HOI4D together with the HOT3D generative model, and estimate only the bandwidth $\hat{\sigma}$\revb{ of Eq.~\ref{eq:deploy_sigma}} from ten unlabeled target clips.}
\rev{Under this calibration-only, no-retraining transfer, StableHand attains the lowest error on every metric on H2O~\cite{kwon2021h2o}\revb{ (Tab.~\ref{tab:supp_transfer}), with $56.77$\,mm W-MPJPE and $30.62$\,s per clip against $60.88$\,mm and $323.89$\,s for} \delb{at about one tenth of the runtime of} the optimization-based Dyn-HaMR~\cite{yu2025dyn}.}
\rev{On HOI4D~\cite{liu2022hoi4d}, articulation and smoothness transfer\revb{, with $8.65$\,mm PA-MPJPE and the lowest AccEr,} while world space positioning does not, which we attribute to its nearly stationary head motion that leaves little parallax for monocular camera estimation.}
\rev{\delb{Full transfer results and the calibration-subset sweep are in the supplemental material.}\revb{Sec.~\ref{supp:transfer} reports the full transfer results, including the calibration-subset sweep in Tab.~\ref{tab:supp_calib_size}.}}

\subsection{Ablation Study}
\label{subsec:exp_ablation}

We ablate StableHand on HOT3D (Tab.~\ref{table:ablation}) along (a)~the quality network, (b)~design choices of the quality-aware flow-matching pathway, and (c)~the quality signal consumed at inference\revb{, and evaluate quality granularity separately on ARCTIC}.

\begin{table}[!ht]
\centering
\caption{\textbf{Ablation studies on HOT3D.}
Three axes: (a)~quality network, (b)~generative model design, and (c)~quality source at inference.
Per-column \colorbox{bestcolor}{\textbf{best}} and \colorbox{secondcolor}{second best} are color-coded within each block.
Per-corpus QN breakdown is in the supplemental material.}
\label{table:ablation}
\resizebox{0.85\linewidth}{!}{
\setlength{\tabcolsep}{2mm}{
\begin{tabular}{lcccc}
\toprule
Setup & PA-MPJPE $\downarrow$ & W-MPJPE $\downarrow$ & WA-MPJPE $\downarrow$ & AccEr $\downarrow$ \\
\midrule
\multicolumn{5}{l}{\emph{(a) Quality network}} \\
\textbf{Ours (full)}                                                         & \best{4.02} & \best{57.83} & \best{21.02} & \second{3.83} \\
QN pretrained on HOT3D only                                                  & 4.85 & \second{61.54} & \second{21.12} & 4.06 \\
QN w/o temporal self-attention (per-frame MLP)                               & \second{4.33} & 64.98 & 25.71 & \best{3.82} \\
\midrule
\multicolumn{5}{l}{\emph{(b) Generative model design}} \\
\textbf{Ours (full)}                                                         & \second{4.02} & \best{57.83} & \best{21.02} & 3.83 \\
w/o quality schedule ($\tilde{\mathbf{q}}{\equiv}\mathbf{0}$, standard FM)   & 4.55 & 92.32 & 30.22 & 4.87 \\
w/o per-component $\mathbf{q}$ (single scalar $q^h$ per hand)                & 5.06 & \second{61.23} & \second{22.37} & 4.89 \\
w/o per-token AdaLN (single global $\mathbf{m}_t$)                           & 4.44 & 64.66 & 23.28 & \best{3.23} \\
w/o smoothness regularizer ($\lambda_{\mathrm{smooth}}{=}0$)                 & 4.62 & 63.57 & 23.26 & \second{3.60} \\
w/o wrist-translation term ($\lambda_{\mathrm{wrist}}{=}0$)                  & \best{3.95} & 70.52 & 22.45 & 4.31 \\
w/o four-token split (single per-frame token)                                & 5.23 & 118.61 & 40.03 & 6.60 \\
\midrule
\multicolumn{5}{l}{\emph{(c) Quality source at inference}} \\
$\mathbf{q}$ = oracle per-component (upper bound)                            & \best{3.22} & \best{48.16} & \best{17.58} & \best{3.12} \\
$\mathbf{q}$ = $\hat{\mathbf{q}}$ from QN (\textbf{Ours (full)})                  & \second{4.02} & \second{57.83} & \second{21.02} & \second{3.83} \\
$\mathbf{q}$ = per-corpus median (constant baseline)                         & 5.43 & 104.62 & 31.37 & 4.25 \\
$\mathbf{q}$ = binary detection mask                                         & 6.19 & 135.35 & 43.82 & 9.39 \\
\bottomrule
\end{tabular}
}}
\end{table}

\noindent\textbf{Quality network (Tab.~\ref{table:ablation}(a)).}
Restricting QN pretraining to HOT3D alone lifts W-MPJPE by $3.7$\,mm to $61.54$, confirming that cross-corpus pretraining broadens failure-mode coverage.
\rev{This cost is small relative to removing the generative design choices below, and broad pretraining alone therefore does not explain the gain.}
Replacing the temporal self-attention encoder with a per-frame MLP further lifts it to $64.98$ ($+7.2$\,mm), confirming that temporal context distinguishes sustained detection gaps from isolated low-confidence frames.

\noindent\textbf{Generative model design (Tab.~\ref{table:ablation}(b)).}
Setting $\tilde{\mathbf{q}}{\equiv}\mathbf{0}$ collapses Eq.~\ref{eq:forward} to a scalar schedule and lifts W-MPJPE to $92.32$ ($1.6\times$): without per-channel quality, the process regenerates every channel from noise instead of anchoring high-quality observations.
Replacing the four tokens per frame with a single per-frame token, which strips AdaLN of any per-channel modulation, is the most damaging entry at $118.61$ ($2.1\times$), confirming that the four-token split is necessary for AdaLN to scale wrist and finger updates separately.
\rev{\delb{Finer channels give diminishing returns: per-finger quality lowers ARCTIC W-MPJPE by a further $3.96$\,mm, while per-joint quality gives no consistent gain over per-finger quality, making the four channels a compact choice rather than an optimum.}}

\noindent\textbf{Quality source at inference (Tab.~\ref{table:ablation}(c)).}
Replacing $\hat{\mathbf{q}}$ with a per-corpus median or a binary detection mask lifts W-MPJPE to $104.62$ and $135.35$, confirming that graded per-component quality cannot be approximated by a constant or detection threshold.
The oracle $\mathbf{q}$ at $48.16$ ($-9.7$\,mm) sits only modestly below our predicted $\hat{\mathbf{q}}$, indicating that the quality network captures most of the available signal.
\rev{Since each row removes a single factor and the factors interact, we read these differences as controlled sensitivities rather than additive shares of the total gain.}
The Wasserstein-loss ablation, RBF bandwidth sensitivity, QN calibration and input-perturbation analyses, \rev{\delb{quality granularity,} quality-network input extensions with texture-free mask statistics and scene depth,} and inference efficiency are in the supplemental material.

{\revbcolor
\noindent\textbf{Quality granularity.}
Separately from the HOT3D ablations of Tab.~\ref{table:ablation}, Tab.~\ref{tab:supp_granularity} evaluates finer quality channels on the $155$ ARCTIC s05 clips.
Per-finger quality lowers W-MPJPE from $124.59$ to $120.63$\,mm, while per-joint quality gives no consistent gain over per-finger quality.
The four channels of the main model (wrist and fingers of each hand) are thus a compact choice rather than an optimum.
\par}

\section{Conclusion}
\label{sec:conclusion}

We presented StableHand, a quality-aware flow-matching framework for world space 4D dual-hand motion estimation, in which a four-value quality signal drives the forward schedule, velocity target, ODE initialization, and AdaLN modulation of a DiT.
StableHand achieves state-of-the-art across every \del{reported metric}\rev{metric of the main comparison\revb{s}} on HOT3D and ARCTIC, with the largest margins on the most occlusion-affected clips.

\noindent\textbf{\rev{Limitations.}}
\delb{The bandwidth of quality signal is calibrated by a specific upstream estimator, and over long low-quality horizons the estimator may drift toward plausible but not faithfully grounded configurations.}\revb{The quality network and its bandwidth are fit to the error distribution of one upstream hand-pose estimator, and over long low-quality horizons StableHand may drift toward plausible but not faithfully grounded configurations.}
\rev{\delb{Quality bandwidth recalibration adapts the error-to-quality scale under a fixed estimator but does not replace retraining after an estimator swap.}\revb{Quality bandwidth recalibration updates no network parameter, and after the WiLoR-to-HaMeR swap it changes W-MPJPE only from $104.81$ to $103.84$\,mm, whereas retraining on HaMeR observations reaches $62.16$\,mm (Sec.~\ref{supp:e2_hamer}).}}
\rev{Calibration-only transfer is shown on egocentric manipulation corpora, and robustness to domains with substantially different visual and motion statistics, such as surgical or laboratory recordings, remains an open question.}
%
%
\revb{World space positioning remains bounded by monocular camera estimation on low-parallax video such as HOI4D, and the objective has no inter-hand collision term, which allows the two hands to interpenetrate.}
\revb{Our metrics are proxies for policy supervision, and end-to-end policy-learning evaluation remains open.}
Joint \rev{object-hand} modeling and end-to-end training are natural next steps.

\begin{ack}
This work is partially supported by ``The MBZUAI-WIS Joint Program for Artificial Intelligence Research''.

\end{ack}

\clearpage
\bibliographystyle{plainnat}
\bibliography{main}

@article{banerjee2024introducing,
  title={Introducing hot3d: An egocentric dataset for 3d hand and object tracking},
  author={Banerjee, Prithviraj and Shkodrani, Sindi and Moulon, Pierre and Hampali, Shreyas and Zhang, Fan and Fountain, Jade and Miller, Edward and Basol, Selen and Newcombe, Richard and Wang, Robert and others},
  journal={arXiv preprint arXiv:2406.09598},
  year={2024}
}

@inproceedings{liu2022hoi4d,
  title={Hoi4d: A 4d egocentric dataset for category-level human-object interaction},
  author={Liu, Yunze and Liu, Yun and Jiang, Che and Lyu, Kangbo and Wan, Weikang and Shen, Hao and Liang, Boqiang and Fu, Zhoujie and Wang, He and Yi, Li},
  booktitle={Proceedings of the IEEE/CVF Conference on Computer Vision and Pattern Recognition},
  pages={21013--21022},
  year={2022}
}

@inproceedings{kwon2021h2o,
  title={H2o: Two hands manipulating objects for first person interaction recognition},
  author={Kwon, Taein and Tekin, Bugra and St{\"u}hmer, Jan and Bogo, Federica and Pollefeys, Marc},
  booktitle={Proceedings of the IEEE/CVF international conference on computer vision},
  pages={10138--10148},
  year={2021}
}

@article{fu2026egograsp,
  title={EgoGrasp: World-Space Hand-Object Interaction Estimation from Egocentric Videos},
  author={Fu, Hongming and Wang, Wenjia and Qiao, Xiaozhen and Potamias, Rolandos Alexandros and Komura, Taku and Yang, Shuo and Liu, Zheng and Zhao, Bo},
  journal={arXiv preprint arXiv:2601.01050},
  year={2026}
}

@inproceedings{ye2023decoupling,
  title={Decoupling human and camera motion from videos in the wild},
  author={Ye, Vickie and Pavlakos, Georgios and Malik, Jitendra and Kanazawa, Angjoo},
  booktitle={Proceedings of the IEEE/CVF conference on computer vision and pattern recognition},
  pages={21222--21232},
  year={2023}
}

@inproceedings{zhang2025hawor,
  title={Hawor: World-space hand motion reconstruction from egocentric videos},
  author={Zhang, Jinglei and Deng, Jiankang and Ma, Chao and Potamias, Rolandos Alexandros},
  booktitle={Proceedings of the Computer Vision and Pattern Recognition Conference},
  pages={1805--1815},
  year={2025}
}

@inproceedings{yu2025dyn,
  title={Dyn-hamr: Recovering 4d interacting hand motion from a dynamic camera},
  author={Yu, Zhengdi and Zafeiriou, Stefanos and Birdal, Tolga},
  booktitle={Proceedings of the Computer Vision and Pattern Recognition Conference},
  pages={27716--27726},
  year={2025}
}

@inproceedings{duran2024hmp,
  title={Hmp: Hand motion priors for pose and shape estimation from video},
  author={Duran, Enes and Kocabas, Muhammed and Choutas, Vasileios and Fan, Zicong and Black, Michael J},
  booktitle={Proceedings of the IEEE/CVF Winter Conference on Applications of Computer Vision},
  pages={6353--6363},
  year={2024}
}

@inproceedings{valassakis2024handdgp,
  title={Handdgp: Camera-space hand mesh prediction with differentiable global positioning},
  author={Valassakis, Eugene and Garcia-Hernando, Guillermo},
  booktitle={European Conference on Computer Vision},
  pages={479--496},
  year={2024},
  organization={Springer}
}

@article{teed2021droid,
  title={Droid-slam: Deep visual slam for monocular, stereo, and rgb-d cameras},
  author={Teed, Zachary and Deng, Jia},
  journal={Advances in neural information processing systems},
  volume={34},
  pages={16558--16569},
  year={2021}
}

@inproceedings{pavlakos2024reconstructing,
  title={Reconstructing hands in 3d with transformers},
  author={Pavlakos, Georgios and Shan, Dandan and Radosavovic, Ilija and Kanazawa, Angjoo and Fouhey, David and Malik, Jitendra},
  booktitle={Proceedings of the IEEE/CVF Conference on Computer Vision and Pattern Recognition},
  pages={9826--9836},
  year={2024}
}

@inproceedings{potamias2025wilor,
  title={Wilor: End-to-end 3d hand localization and reconstruction in-the-wild},
  author={Potamias, Rolandos Alexandros and Zhang, Jinglei and Deng, Jiankang and Zafeiriou, Stefanos},
  booktitle={Proceedings of the Computer Vision and Pattern Recognition Conference},
  pages={12242--12254},
  year={2025}
}

@inproceedings{lin2021mesh,
  title={Mesh graphormer},
  author={Lin, Kevin and Wang, Lijuan and Liu, Zicheng},
  booktitle={Proceedings of the IEEE/CVF international conference on computer vision},
  pages={12939--12948},
  year={2021}
}

@inproceedings{liu2021semi,
  title={Semi-supervised 3d hand-object poses estimation with interactions in time},
  author={Liu, Shaowei and Jiang, Hanwen and Xu, Jiarui and Liu, Sifei and Wang, Xiaolong},
  booktitle={Proceedings of the IEEE/CVF conference on computer vision and pattern recognition},
  pages={14687--14697},
  year={2021}
}

@inproceedings{park2022handoccnet,
  title={Handoccnet: Occlusion-robust 3d hand mesh estimation network},
  author={Park, JoonKyu and Oh, Yeonguk and Moon, Gyeongsik and Choi, Hongsuk and Lee, Kyoung Mu},
  booktitle={Proceedings of the IEEE/CVF conference on computer vision and pattern recognition},
  pages={1496--1505},
  year={2022}
}

@inproceedings{fu2023deformer,
  title={Deformer: Dynamic fusion transformer for robust hand pose estimation},
  author={Fu, Qichen and Liu, Xingyu and Xu, Ran and Niebles, Juan Carlos and Kitani, Kris M},
  booktitle={Proceedings of the IEEE/CVF International Conference on Computer Vision},
  pages={23600--23611},
  year={2023}
}

@article{sun2026unihand,
  title={UniHand: A Unified Model for Diverse Controlled 4D Hand Motion Modeling},
  author={Sun, Zhihao and Wu, Tong and Tu, Ruirui and Dong, Daoguo and Wu, Zuxuan},
  journal={arXiv preprint arXiv:2602.21631},
  year={2026}
}

@inproceedings{zhou2019continuity,
  title={On the continuity of rotation representations in neural networks},
  author={Zhou, Yi and Barnes, Connelly and Lu, Jingwan and Yang, Jimei and Li, Hao},
  booktitle={Proceedings of the IEEE/CVF conference on computer vision and pattern recognition},
  pages={5745--5753},
  year={2019}
}

@article{MANO:SIGGRAPHASIA:2017,
      title = {Embodied Hands: Modeling and Capturing Hands and Bodies Together},
      author = {Romero, Javier and Tzionas, Dimitrios and Black, Michael J.},
      journal = {ACM Transactions on Graphics, (Proc. SIGGRAPH Asia)},
      volume = {36},
      number = {6},
      series = {245:1--245:17},
      month = nov,
      year = {2017},
      month_numeric = {11}
  }

@article{ionescu2013human3,
  title={Human3. 6m: Large scale datasets and predictive methods for 3d human sensing in natural environments},
  author={Ionescu, Catalin and Papava, Dragos and Olaru, Vlad and Sminchisescu, Cristian},
  journal={IEEE transactions on pattern analysis and machine intelligence},
  volume={36},
  number={7},
  pages={1325--1339},
  year={2013},
  publisher={IEEE}
}

@article{lipman2022flow,
  title={Flow matching for generative modeling},
  author={Lipman, Yaron and Chen, Ricky TQ and Ben-Hamu, Heli and Nickel, Maximilian and Le, Matt},
  journal={arXiv preprint arXiv:2210.02747},
  year={2022}
}

@inproceedings{peebles2023scalable,
  title={Scalable diffusion models with transformers},
  author={Peebles, William and Xie, Saining},
  booktitle={Proceedings of the IEEE/CVF international conference on computer vision},
  pages={4195--4205},
  year={2023}
}

@article{su2024roformer,
  title={Roformer: Enhanced transformer with rotary position embedding},
  author={Su, Jianlin and Ahmed, Murtadha and Lu, Yu and Pan, Shengfeng and Bo, Wen and Liu, Yunfeng},
  journal={Neurocomputing},
  volume={568},
  pages={127063},
  year={2024},
  publisher={Elsevier}
}

@article{lin2025depth,
  title={Depth anything 3: Recovering the visual space from any views},
  author={Lin, Haotong and Chen, Sili and Liew, Junhao and Chen, Donny Y and Li, Zhenyu and Shi, Guang and Feng, Jiashi and Kang, Bingyi},
  journal={arXiv preprint arXiv:2511.10647},
  year={2025}
}

@article{meng2021sdedit,
  title={Sdedit: Guided image synthesis and editing with stochastic differential equations},
  author={Meng, Chenlin and He, Yutong and Song, Yang and Song, Jiaming and Wu, Jiajun and Zhu, Jun-Yan and Ermon, Stefano},
  journal={arXiv preprint arXiv:2108.01073},
  year={2021}
}

@article{vaswani2017attention,
  title={Attention is all you need},
  author={Vaswani, Ashish and Shazeer, Noam and Parmar, Niki and Uszkoreit, Jakob and Jones, Llion and Gomez, Aidan N and Kaiser, {\L}ukasz and Polosukhin, Illia},
  journal={Advances in neural information processing systems},
  volume={30},
  year={2017}
}

@inproceedings{moon2020interhand2,
  title={Interhand2. 6m: A dataset and baseline for 3d interacting hand pose estimation from a single rgb image},
  author={Moon, Gyeongsik and Yu, Shoou-I and Wen, He and Shiratori, Takaaki and Lee, Kyoung Mu},
  booktitle={European Conference on Computer Vision},
  pages={548--564},
  year={2020},
  organization={Springer}
}

@article{moon2023dataset,
  title={A dataset of relighted 3D interacting hands},
  author={Moon, Gyeongsik and Saito, Shunsuke and Xu, Weipeng and Joshi, Rohan and Buffalini, Julia and Bellan, Harley and Rosen, Nicholas and Richardson, Jesse and Mize, Mallorie and De Bree, Philippe and others},
  journal={Advances in Neural Information Processing Systems},
  volume={36},
  pages={17689--17701},
  year={2023}
}

@inproceedings{fan2023arctic,
  title={ARCTIC: A dataset for dexterous bimanual hand-object manipulation},
  author={Fan, Zicong and Taheri, Omid and Tzionas, Dimitrios and Kocabas, Muhammed and Kaufmann, Manuel and Black, Michael J and Hilliges, Otmar},
  booktitle={Proceedings of the IEEE/CVF conference on computer vision and pattern recognition},
  pages={12943--12954},
  year={2023}
}

@article{hoque2025egodex,
  title={Egodex: Learning dexterous manipulation from large-scale egocentric video},
  author={Hoque, Ryan and Huang, Peide and Yoon, David J and Sivapurapu, Mouli and Zhang, Jian},
  journal={arXiv preprint arXiv:2505.11709},
  year={2025}
}

@article{punamiya2026egoverse,
  title={EgoVerse: An Egocentric Human Dataset for Robot Learning from Around the World},
  author={Punamiya, Ryan and Kareer, Simar and Liu, Zeyi and Citron, Josh and Qiu, Ri-Zhao and Cai, Xiongyi and Gavryushin, Alexey and Chen, Jiaqi and Liconti, Davide and Zhu, Lawrence Y and others},
  journal={arXiv preprint arXiv:2604.07607},
  year={2026}
}

@inproceedings{grauman2024ego,
  title={Ego-exo4d: Understanding skilled human activity from first-and third-person perspectives},
  author={Grauman, Kristen and Westbury, Andrew and Torresani, Lorenzo and Kitani, Kris and Malik, Jitendra and Afouras, Triantafyllos and Ashutosh, Kumar and Baiyya, Vijay and Bansal, Siddhant and Boote, Bikram and others},
  booktitle={Proceedings of the IEEE/CVF Conference on Computer Vision and Pattern Recognition},
  pages={19383--19400},
  year={2024}
}

@inproceedings{perrett2025hd,
  title={Hd-epic: A highly-detailed egocentric video dataset},
  author={Perrett, Toby and Darkhalil, Ahmad and Sinha, Saptarshi and Emara, Omar and Pollard, Sam and Parida, Kranti Kumar and Liu, Kaiting and Gatti, Prajwal and Bansal, Siddhant and Flanagan, Kevin and others},
  booktitle={Proceedings of the Computer Vision and Pattern Recognition Conference},
  pages={23901--23913},
  year={2025}
}

@article{yang2025egovla,
  title={Egovla: Learning vision-language-action models from egocentric human videos},
  author={Yang, Ruihan and Yu, Qinxi and Wu, Yecheng and Yan, Rui and Li, Borui and Cheng, An-Chieh and Zou, Xueyan and Fang, Yunhao and Cheng, Xuxin and Qiu, Ri-Zhao and others},
  journal={arXiv preprint arXiv:2507.12440},
  year={2025}
}

@inproceedings{li2025maniptrans,
  title={Maniptrans: Efficient dexterous bimanual manipulation transfer via residual learning},
  author={Li, Kailin and Li, Puhao and Liu, Tengyu and Li, Yuyang and Huang, Siyuan},
  booktitle={Proceedings of the IEEE/CVF Conference on Computer Vision and Pattern Recognition},
  pages={6991--7003},
  year={2025}
}

@inproceedings{prakash20243d,
  title={3d hand pose estimation in everyday egocentric images},
  author={Prakash, Aditya and Tu, Ruisen and Chang, Matthew and Gupta, Saurabh},
  booktitle={European Conference on Computer Vision},
  pages={183--202},
  year={2024},
  organization={Springer}
}

@inproceedings{cao2021reconstructing,
  title={Reconstructing hand-object interactions in the wild},
  author={Cao, Zhe and Radosavovic, Ilija and Kanazawa, Angjoo and Malik, Jitendra},
  booktitle={Proceedings of the IEEE/CVF international conference on computer vision},
  pages={12417--12426},
  year={2021}
}

@inproceedings{kareer2025egomimic,
  title={Egomimic: Scaling imitation learning via egocentric video},
  author={Kareer, Simar and Patel, Dhruv and Punamiya, Ryan and Mathur, Pranay and Cheng, Shuo and Wang, Chen and Hoffman, Judy and Xu, Danfei},
  booktitle={2025 IEEE International Conference on Robotics and Automation (ICRA)},
  pages={13226--13233},
  year={2025},
  organization={IEEE}
}

@article{zhu2026emma,
  title={Emma: Scaling mobile manipulation via egocentric human data},
  author={Zhu, Lawrence Y and Kuppili, Pranav and Punamiya, Ryan and Aphiwetsa, Patcharapong and Patel, Dhruv and Kareer, Simar and Ha, Sehoon and Xu, Danfei},
  journal={IEEE Robotics and Automation Letters},
  year={2026},
  publisher={IEEE}
}

@article{zheng2026egoscale,
  title={Egoscale: Scaling dexterous manipulation with diverse egocentric human data},
  author={Zheng, Ruijie and Niu, Dantong and Xie, Yuqi and Wang, Jing and Xu, Mengda and Jiang, Yunfan and Casta{\~n}eda, Fernando and Hu, Fengyuan and Tan, You Liang and Fu, Letian and others},
  journal={arXiv preprint arXiv:2602.16710},
  year={2026}
}

@article{li2025scalable,
  title={Scalable vision-language-action model pretraining for robotic manipulation with real-life human activity videos},
  author={Li, Qixiu and Deng, Yu and Liang, Yaobo and Luo, Lin and Zhou, Lei and Yao, Chengtang and Zeng, Lingqi and Feng, Zhiyuan and Liang, Huizhi and Xu, Sicheng and others},
  journal={arXiv preprint arXiv:2510.21571},
  year={2025}
}

@article{guzey2025dexterity,
  title={Dexterity from Smart Lenses: Multi-Fingered Robot Manipulation with In-the-Wild Human Demonstrations},
  author={Guzey, Irmak and Qi, Haozhi and Urain, Julen and Wang, Changhao and Yin, Jessica and Bodduluri, Krishna and Lambeta, Mike and Pinto, Lerrel and Rai, Akshara and Malik, Jitendra and others},
  journal={arXiv preprint arXiv:2511.16661},
  year={2025}
}

@inproceedings{lugmayr2022repaint,
  title={Repaint: Inpainting using denoising diffusion probabilistic models},
  author={Lugmayr, Andreas and Danelljan, Martin and Romero, Andres and Yu, Fisher and Timofte, Radu and Van Gool, Luc},
  booktitle={Proceedings of the IEEE/CVF conference on computer vision and pattern recognition},
  pages={11461--11471},
  year={2022}
}

@article{sahoo2024diffusion,
  title={Diffusion models with learned adaptive noise},
  author={Sahoo, Subham S and Gokaslan, Aaron and De, Chris and Kuleshov, Volodymyr},
  journal={Advances in Neural Information Processing Systems},
  volume={37},
  pages={105730--105779},
  year={2024}
}

@inproceedings{boukhayma20193d,
  title={3d hand shape and pose from images in the wild},
  author={Boukhayma, Adnane and Bem, Rodrigo de and Torr, Philip HS},
  booktitle={Proceedings of the IEEE/CVF conference on computer vision and pattern recognition},
  pages={10843--10852},
  year={2019}
}

@inproceedings{lin2021end,
  title={End-to-end human pose and mesh reconstruction with transformers},
  author={Lin, Kevin and Wang, Lijuan and Liu, Zicheng},
  booktitle={Proceedings of the IEEE/CVF conference on computer vision and pattern recognition},
  pages={1954--1963},
  year={2021}
}

@article{ho2020denoising,
  title={Denoising diffusion probabilistic models},
  author={Ho, Jonathan and Jain, Ajay and Abbeel, Pieter},
  journal={Advances in neural information processing systems},
  volume={33},
  pages={6840--6851},
  year={2020}
}

@article{zeng2026flowhoi,
  title={FlowHOI: Flow-based Semantics-Grounded Generation of Hand-Object Interactions for Dexterous Robot Manipulation},
  author={Zeng, Huajian and Chen, Lingyun and Yang, Jiaqi and Zhang, Yuantai and Shi, Fan and Liu, Peidong and Zuo, Xingxing},
  journal={arXiv preprint arXiv:2602.13444},
  year={2026}
}

@inproceedings{mu2025stablemotion,
  title={StableMotion: Training Motion Cleanup Models with Unpaired Corrupted Data},
  author={Mu, Yuxuan and Ling, Hung Yu and Shi, Yi and Ojeda, Ismael Baira and Xi, Pengcheng and Shu, Chang and Zinno, Fabio and Peng, Xue Bin},
  booktitle={Proceedings of the SIGGRAPH Asia 2025 Conference Papers},
  pages={1--12},
  year={2025}
}

@book{ambrosio2005gradient,
  title={Gradient flows: in metric spaces and in the space of probability measures},
  author={Ambrosio, Luigi and Gigli, Nicola and Savar{\'e}, Giuseppe},
  year={2005},
  publisher={Springer}
}

@article{ye2026whole,
  title={Whole: World-grounded hand-object lifted from egocentric videos},
  author={Ye, Yufei and Li, Jiaman and Rong, Ryan and Liu, C Karen},
  journal={arXiv preprint arXiv:2602.22209},
  year={2026}
}

@book{scholkopf2002learning,
  title={Learning with kernels: support vector machines, regularization, optimization, and beyond},
  author={Sch{\"o}lkopf, Bernhard and Smola, Alexander J},
  publisher={MIT Press},
  year={2002}
}

@article{loshchilov2017decoupled,
  title={Decoupled weight decay regularization},
  author={Loshchilov, Ilya and Hutter, Frank},
  journal={arXiv preprint arXiv:1711.05101},
  year={2017}
}

@article{shazeer2020glu,
  title={Glu variants improve transformer},
  author={Shazeer, Noam},
  journal={arXiv preprint arXiv:2002.05202},
  year={2020}
}

@article{chi2024estimating,
  title={Estimating ego-body pose from doubly sparse egocentric video data},
  author={Chi, Seunggeun and Huang, Pin-Hao and Sachdeva, Enna and Ma, Hengbo and Ramani, Karthik and Lee, Kwonjoon},
  journal={Advances in neural information processing systems},
  volume={37},
  pages={55178--55203},
  year={2024}
}

@article{hatano2025invisible,
  title={The invisible egohand: 3d hand forecasting through egobody pose estimation},
  author={Hatano, Masashi and Zhu, Zhifan and Saito, Hideo and Damen, Dima},
  journal={arXiv preprint arXiv:2504.08654},
  year={2025}
}

@inproceedings{ye2026predicting,
  title={Predicting 4d hand trajectory from monocular videos},
  author={Ye, Yufei and Feng, Yao and Taheri, Omid and Feng, Haiwen and Tulsiani, Shubham and Black, Michael J},
  booktitle={2026 International Conference on 3D Vision (3DV)},
  pages={860--870},
  year={2026},
  organization={IEEE}
}

@article{saroha2026egoflow,
  title={EgoFlow: Gradient-Guided Flow Matching for Egocentric 6DoF Object Motion Generation},
  author={Saroha, Abhishek and Zeng, Huajian and Zuo, Xingxing and Cremers, Daniel and Wang, Xi},
  journal={arXiv preprint arXiv:2604.01421},
  year={2026}
}

\clearpage
\appendix
\clearpage

\begin{center}
  {\LARGE\bfseries Supplementary Material}
\end{center}
\bigskip

This supplementary document provides additional details and results that complement the main paper.
Sec.~\ref{supp:implementation} summarizes the full architecture and training hyperparameters.
Sec.~\ref{supp:additional_analyses} reports additional robustness and calibration analyses spanning predicted-quality calibration against ground truth, quality-network input perturbations, and estimator generality across upstream pose estimators.
\rev{Sec.~\ref{supp:transfer} reports calibration-only transfer to H2O and HOI4D.}
Sec.~\ref{supp:ablation} reports supplemental ablation studies covering the radial-basis bandwidth (Sec.~\ref{supp:sigma_sweep})\rev{,} \del{and} the quality network architecture (Sec.~\ref{supp:qn_ablation})\rev{, quality-network input extensions (Sec.~\ref{supp:qn_extensions}), and quality granularity (Sec.~\ref{supp:granularity})}.
Sec.~\ref{supp:runtime} reports inference efficiency including a per-stage runtime breakdown and an ODE-step sweep.
Sec.~\ref{supp:augmentation} specifies the synthetic perturbation taxonomy used to augment the quality-network training distribution.
Sec.~\ref{supp:datasets} details the eight egocentric corpora used to pretrain the quality network.
Sec.~\ref{supp:metrics} provides formal definitions of the evaluation metrics.
Sec.~\ref{supp:more_qualitative} \del{concludes with}\rev{reports} additional qualitative results and failure-case analyses\rev{, and Sec.~\ref{supp:in_the_wild} concludes with a qualitative illustration of in-the-wild transfer}.

\section{Implementation Details}
\label{supp:implementation}

The DiT denoiser stacks $8$ blocks of hidden dimension $512$ with $8$ attention heads and a $4{\times}$ GeGLU~\cite{shazeer2020glu} feed-forward expansion.
Each frame is tokenized into four tokens (left/right wrist, left/right fingers) and temporal RoPE~\cite{su2024roformer} self-attention spans all $4T$ tokens.
The quality network is a $4$-layer Transformer encoder of hidden dimension $256$ and $4$ attention heads with camera-state cross-attention, and shares weights between the two hands.
The total parameter count is $70.7$M (DiT $66.8$M + QN $3.6$M + auxiliary heads $0.3$M).

Both networks are trained on a single NVIDIA H100 GPU under bfloat16 autocast with AdamW~\cite{loshchilov2017decoupled} (learning rate $1{\times}10^{-4}$, weight decay $0.01$, batch size $16$, $1000$-step linear warmup followed by cosine decay, gradient clipping $\|\nabla\|{\leq}1.0$, seed $42$).
A single benchmark reaches early-stop on \emph{eval/W-MPJPE} in $6$--$8$ H100-hours.
The $20$-step Euler ODE solve through the DiT denoiser consumes ${\sim}230$\,ms per clip.
The full pipeline including upstream estimators is detailed in Sec.~\ref{supp:runtime}.
The pose estimator and geometry model are the released WiLoR~\cite{potamias2025wilor} and Depth-Anything-V3~\cite{lin2025depth} checkpoints, with no fine-tuning.

\section{Additional Quality-Network and Estimator Analyses}
\label{supp:additional_analyses}

This section reports three robustness and calibration analyses that complement the ablation study of the main paper.
Sec.~\ref{supp:e8_calibration} reports a per-channel calibration analysis of the predicted quality signal against ground truth.
Sec.~\ref{supp:e1_perturbation} probes the quality network's response to controlled noise on each of its input streams and the corresponding downstream impact.
Sec.~\ref{supp:e2_hamer} examines whether the framework transfers to a different upstream pose estimator (HaMeR).
Sec.~\ref{supp:e1_perturbation} and Sec.~\ref{supp:e2_hamer} report the 16-joint MANO MPJPE for direct comparison with Tab.~\ref{table:hot3d} of the main paper, while Sec.~\ref{supp:e8_calibration} color-codes its scatter by per-frame wrist-joint W-MPJPE for visual correspondence with the wrist quality channel.

\subsection{Predicted-Quality Calibration}
\label{supp:e8_calibration}

\begin{figure}[h]
  \centering
  \includegraphics[width=0.95\linewidth]{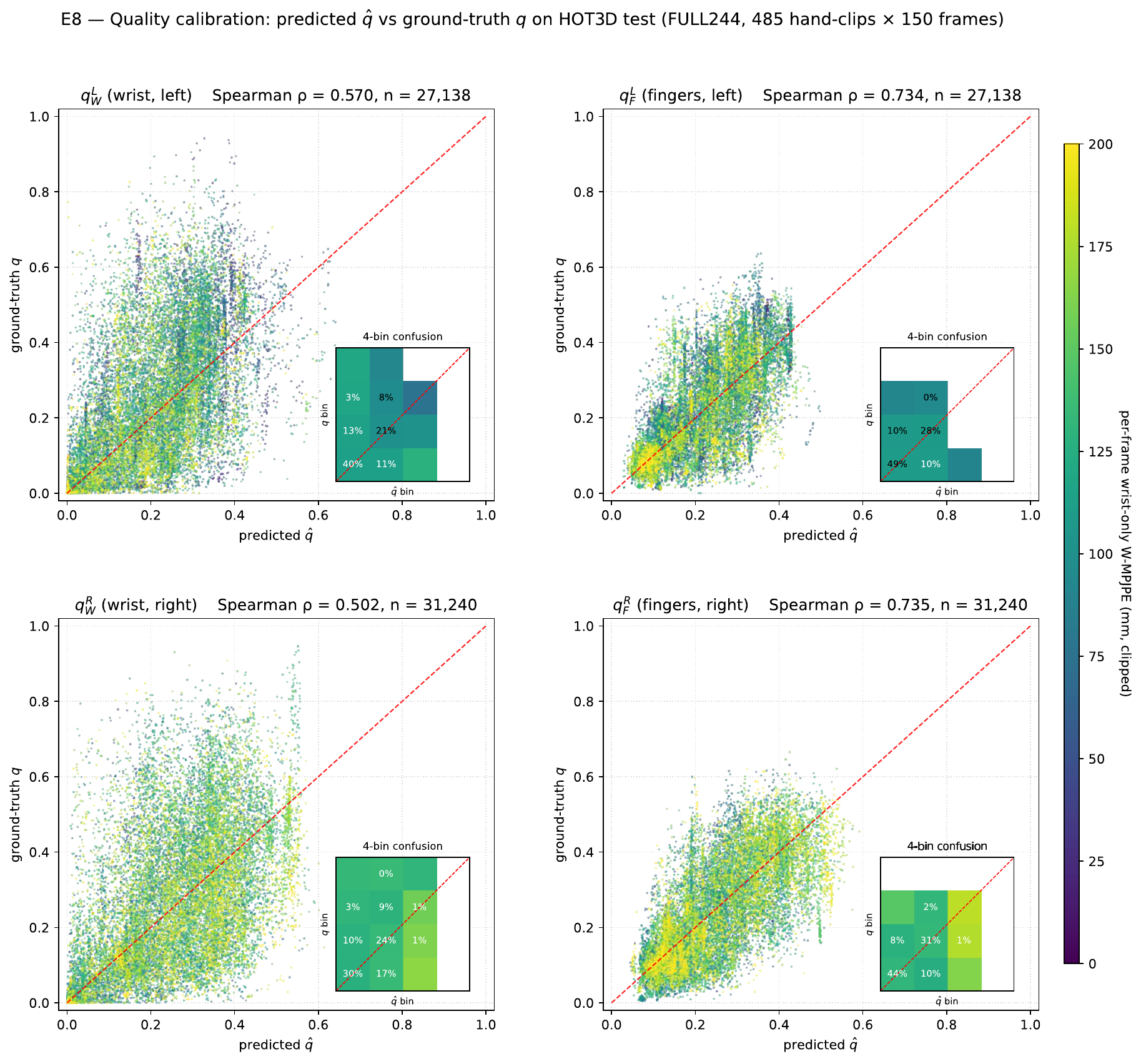}
  \caption{\textbf{Predicted-quality calibration on HOT3D test.}
  Each panel corresponds to one of the four quality channels.
  Each scatter point is one (frame, hand) pair, colored by the per-frame wrist-joint W-MPJPE (viridis colormap).
  The diagonal line marks perfect calibration, and the inset $4{\times}4$ matrix quantizes both axes into four bins (V-Low, Low, Mid, High) with cell percentages.
  Spearman rank correlations are reported per channel.}
  \label{fig:supp_e8_calibration}
\end{figure}

Fig.~\ref{fig:supp_e8_calibration} reports a per-channel calibration analysis of the predicted quality signal against ground truth on the HOT3D test split, since the QN is the central conditioning signal of the generative model and its calibration directly bounds downstream behavior.
For each of the four quality channels ($q^L_W, q^L_F, q^R_W, q^R_F$), we plot a scatter of (predicted, ground-truth) pairs across all valid frames, color-coded by the per-frame downstream wrist-joint W-MPJPE, with an inset $4{\times}4$ confusion matrix that quantizes both axes into four bins.
The Spearman rank correlation $\rho(\hat{\mathbf{q}}, \mathbf{q})$ is $0.734$ for the left finger channel, $0.735$ for the right finger channel, $0.570$ for the left wrist channel, and $0.502$ for the right wrist channel.
Wrist channels are systematically harder to calibrate because wrist global trajectory error involves both translation and rotation drift in the world frame, while finger error is confined to the local articulation manifold.

Off-diagonal frames in the four-bin grid (where the predicted-q bin differs from the ground-truth bin) account for $22\%$ to $44\%$ of frames depending on the channel.
The mean downstream wrist-joint W-MPJPE on these off-diagonal frames is within $2$ to $3$\,mm of the on-diagonal mean, indicating that QN miscalibration is bounded in its downstream impact: even when the QN occasionally over- or under-confidently classifies a frame, the generative process degrades gracefully rather than collapsing to the prior or anchoring to a noisy observation.

\subsection{Quality-Network Input Perturbation}
\label{supp:e1_perturbation}

We probe the quality network's downstream value by injecting controlled noise into each of its input streams and measuring both the QN-side error degradation and the wrist trajectory response.
Each of the three QN inputs (the per-hand observation $\bar{\mathbf{y}}$, the binary detection flag $\delta$, and the per-frame camera pose $\mathbf{g}_t$) is perturbed at four severity levels covering deployment-realistic to extreme noise, with all other streams held clean.
Inference uses the frozen quality network and frozen flow-matching denoiser without retraining, recording the change in predicted error $\Delta\hat{\mathbf{e}}$ and the change in downstream W-MPJPE relative to the unperturbed baseline, which matches the corresponding row of Tab.~\ref{table:hot3d} to within rounding.
For context, we further reference the oracle-q upper bound and constant-q lower bound from Tab.~\ref{table:ablation}(c), obtained from separately trained DiTs using ground-truth and per-corpus-median quality at both training and inference.

\begin{table}[h]
\centering
\caption{\textbf{Quality-network input perturbation sweep on HOT3D test.}
$\bar{\hat{e}}$ denotes the mean predicted error magnitude across all components, and W-MPJPE is the downstream world space MPJPE under the 16-joint MANO MPJPE convention of Tab.~\ref{table:hot3d}.
$\Delta$ columns report the change relative to the predicted-q baseline at zero perturbation.
\protect\rev{The $\mathbf{g}_t$ rows perturb the camera-pose input of the quality network, not the hand observation $\bar{\mathbf{y}}$.}}
\label{tab:supp_e1_perturbation}
\small
\setlength{\tabcolsep}{6pt}
\renewcommand{\arraystretch}{1.05}
\begin{tabular}{@{}llrrrr@{}}
\toprule
\textbf{Stream} & \textbf{Level} & \multicolumn{1}{c}{$\bar{\hat{e}}$ (mm)} & \multicolumn{1}{c}{$\Delta\hat{e}$ (mm)} & \multicolumn{1}{c}{W-MPJPE (mm)} & \multicolumn{1}{c}{$\Delta W$ (mm)} \\
\midrule
\rowcolor{black!4}
\multicolumn{6}{@{}l}{\itshape Reference anchors from Tab.~\ref{table:ablation}(c) (separately trained DiTs)} \\
\rowcolor{black!4}
\quad oracle-q (upper bound) & --- & --- & --- & 48.16 & --- \\
\rowcolor{black!4}
\quad constant-q (lower bound) & --- & --- & --- & 104.62 & --- \\
\midrule
\multicolumn{6}{@{}l}{\itshape Predicted-q (Ours) baseline + perturbation sweep} \\
\textbf{baseline} (no perturbation) & 0 & 27.36 & +0.00 & \textbf{57.83} & +0.00 \\
\cmidrule(lr){1-6}
\multirow{4}{*}{$\bar{\mathbf{y}}$ trans (mm)}
 & 1  & 27.36 & $-0.00$ & 57.06 & $-0.77$ \\
 & 5  & 27.36 & $+0.01$ & 57.17 & $-0.66$ \\
 & 10 & 27.39 & $+0.03$ & 57.58 & $-0.25$ \\
 & 30 & 27.66 & $+0.30$ & 59.62 & $+1.79$ \\
\cmidrule(lr){1-6}
\multirow{4}{*}{$\bar{\mathbf{y}}$ rot (deg)}
 & 0.5 & 27.35 & $-0.00$ & 58.56 & $+0.73$ \\
 & 1   & 27.36 & $-0.00$ & 57.20 & $-0.63$ \\
 & 2   & 27.37 & $+0.01$ & 57.45 & $-0.38$ \\
 & 5   & 27.46 & $+0.10$ & 57.57 & $-0.26$ \\
\cmidrule(lr){1-6}
\multirow{4}{*}{$\bar{\mathbf{y}}$ AA (rad)}
 & 0.01 & 27.39 & $+0.03$ & 57.64 & $-0.19$ \\
 & 0.05 & 28.14 & $+0.79$ & 58.57 & $+0.74$ \\
 & 0.1  & 30.08 & $+2.72$ & 65.55 & $+7.72$ \\
 & 0.2  & 35.35 & $+8.00$ & \textbf{99.15} & \textbf{$+41.32$} \\
\cmidrule(lr){1-6}
\multirow{4}{*}{$\delta$ flip rate}
 & 0.05 & 27.79 & $+0.43$ & 57.67 & $-0.16$ \\
 & 0.1  & 28.27 & $+0.92$ & 57.98 & $+0.15$ \\
 & 0.2  & 29.09 & $+1.73$ & 58.22 & $+0.39$ \\
 & 0.5  & 31.08 & $+3.72$ & 60.90 & $+3.07$ \\
\cmidrule(lr){1-6}
\multirow{4}{*}{$\mathbf{g}_t$ trans (mm)}
 & 1  & 27.36 & $-0.00$ & 57.77 & $-0.06$ \\
 & 5  & 27.36 & $+0.00$ & 57.67 & $-0.16$ \\
 & 10 & 27.36 & $+0.00$ & 57.02 & $-0.81$ \\
 & 30 & 27.40 & $+0.04$ & 59.64 & $+1.81$ \\
\cmidrule(lr){1-6}
\multirow{4}{*}{$\mathbf{g}_t$ rot (deg)}
 & 0.5 & 27.36 & $+0.00$ & 56.88 & $-0.95$ \\
 & 1   & 27.36 & $+0.01$ & 56.89 & $-0.94$ \\
 & 2   & 27.39 & $+0.04$ & 57.56 & $-0.27$ \\
 & 5   & 27.59 & $+0.24$ & \textbf{63.34} & \textbf{$+5.51$} \\
\midrule
combined moderate & all level-2 & 29.19 & $+1.84$ & 59.38 & $+1.55$ \\
\bottomrule
\end{tabular}

\end{table}

\begin{figure}[h]
  \centering
  \includegraphics[width=0.95\linewidth]{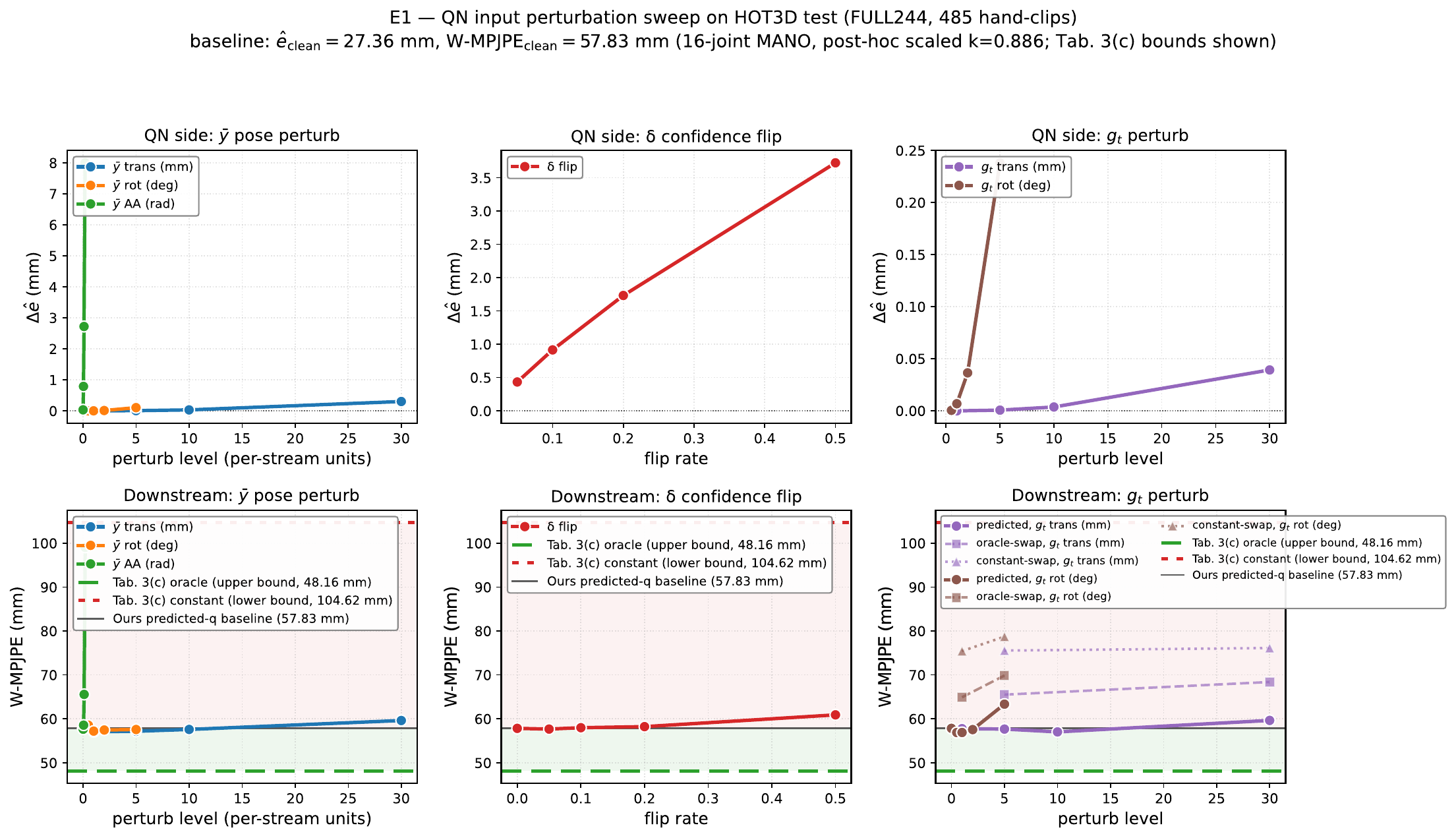}
  \caption{\textbf{Quality-network input perturbation sweep.}
  Top row: change in predicted error $\Delta\hat{\mathbf{e}}$ (mm) under controlled perturbation of each QN input stream; bottom row: downstream W-MPJPE (mm) under the same perturbations, with horizontal references for the predicted-q baseline ($57.83$\,mm), the oracle-q upper bound ($48.16$\,mm), and the constant-q lower bound ($104.62$\,mm) from Tab.~\ref{table:ablation}(c).
  Columns correspond to the per-hand observation $\bar{\mathbf{y}}$, the detection flag $\delta$, and the camera pose $\mathbf{g}_t$, with sub-streams plotted within each column.
  The $\mathbf{g}_t$ panel additionally overlays oracle-q and constant-q reference curves to disentangle DiT-side from QN-side bottlenecks under camera-pose noise.
  See Tab.~\ref{tab:supp_e1_perturbation} for the full numerical sweep.}
  \label{fig:supp_e1_perturbation}
\end{figure}

Tab.~\ref{tab:supp_e1_perturbation} reports the full sweep, and Fig.~\ref{fig:supp_e1_perturbation} summarizes the same data graphically.
At zero perturbation the predicted-q baseline of $57.83$\,mm sits between the oracle-q upper bound of $48.16$\,mm and the constant-q lower bound of $104.62$\,mm from Tab.~\ref{table:ablation}(c).
This recovers $83\%$ of the gap an explicit per-frame quality signal can close relative to a constant-q ablation, leaving a $9.7$\,mm residual to the oracle that reflects QN prediction error rather than an architectural limitation of the quality-aware pathway.

Across all but the most extreme perturbation level of each input stream in Tab.~\ref{tab:supp_e1_perturbation}, downstream W-MPJPE stays within $1$\,mm of the unperturbed baseline.

At extreme noise the largest sensitivity is finger axis-angle perturbation at $0.2$\,rad ($+41.3$\,mm), which scrambles articulation that the DiT cannot recover.
Camera rotation at $5^\circ$ ($+5.5$\,mm) feeds the DiT cross-attention directly and similarly degrades the inference-swap oracle-q reference line in Fig.~\ref{fig:supp_e1_perturbation}, indicating a DiT-side bottleneck rather than a QN failure.
This experiment subsumes the controllably-corrupted-$\hat{\mathbf{q}}$ probe, since every realistic $\hat{\mathbf{q}}$ corruption arises from upstream input noise and the input-side perturbation curves bound it from above.

\rev{A $30$\,mm translation of $\bar{\mathbf{y}}$ raises $\bar{\hat{e}}$ by only $0.30$\,mm, against $8.0$\,mm for a $0.2$\,rad articulation perturbation.}
\rev{The quality network thus responds weakly to a smooth global hand translation, which motivates the scene-depth input of Sec.~\ref{supp:qn_extensions}.}

\subsection{Estimator Generality on HaMeR Observations}
\label{supp:e2_hamer}

The framework predicts the per-component error in physical units (mm) from a generic MANO observation, a binary detection flag, and a head-camera pose, none of which are estimator-specific, so the quality-aware pipeline transfers in principle to any pose estimator that produces MANO predictions.
We empirically probe this generality by comparing the per-component error distributions of WiLoR and HaMeR on the same HOT3D test data, by running drop-in inference with HaMeR observations under two calibration settings (\del{zero-shot}\rev{uncalibrated} and bandwidth-recalibrated), and by retraining the entire pipeline end-to-end on HaMeR observations.

\begin{figure}[h]
  \centering
  \includegraphics[width=0.85\linewidth]{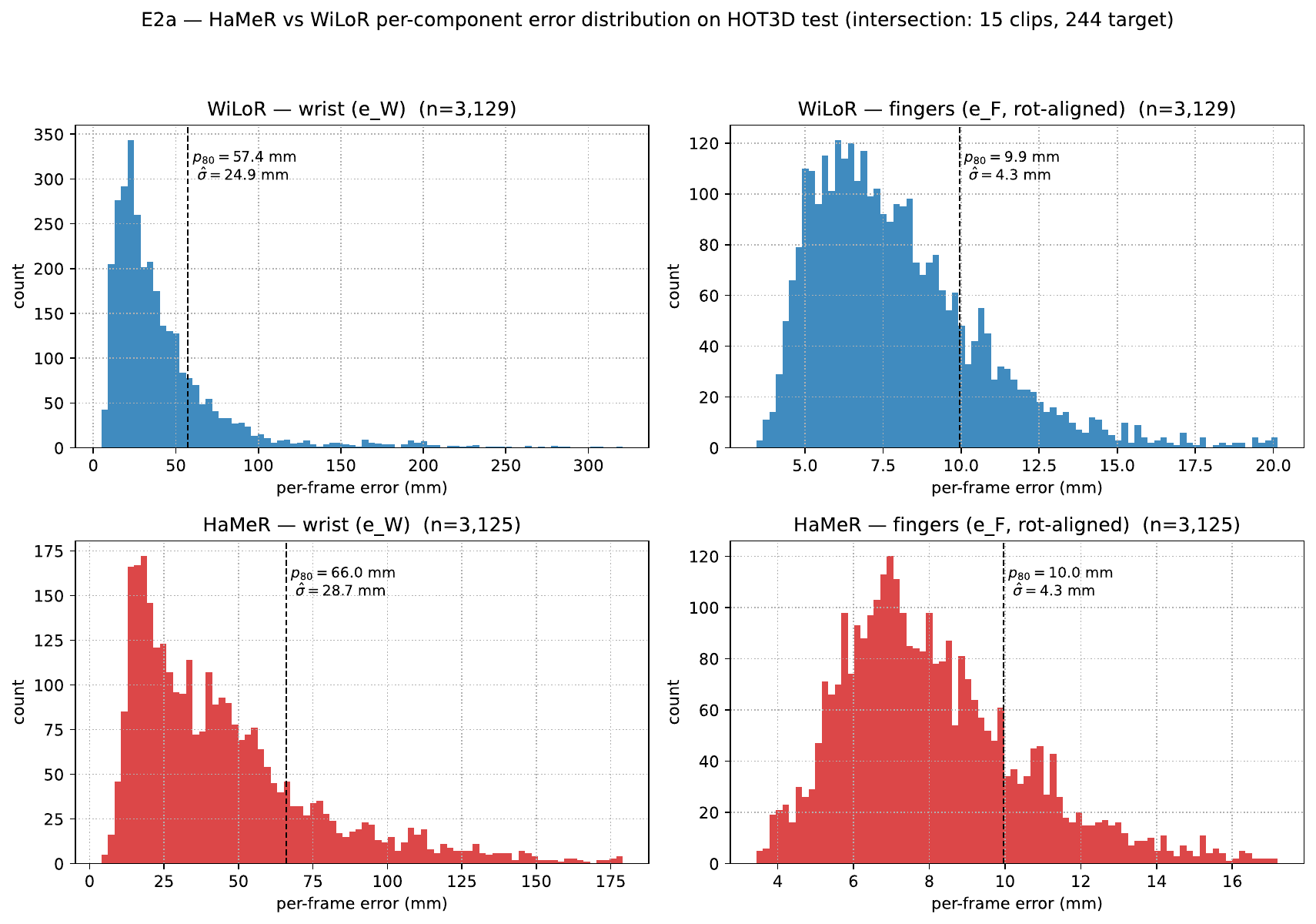}
  \caption{\textbf{Per-component error distributions on HOT3D under two upstream pose estimators.}
  Histograms of $e_W$ (wrist error) and $e_F$ (finger MPJPE in the wrist frame) for WiLoR and HaMeR, computed on the same $15$ HOT3D test clips with both estimators cached.
  Vertical lines mark the $80$th-percentile error per estimator.}
  \label{fig:supp_e2_distribution}
\end{figure}

Fig.~\ref{fig:supp_e2_distribution} overlays the per-component error histograms produced by the two estimators on $15$ HOT3D test clips with both estimator caches available, totaling roughly $3{,}125$ wrist samples and $3{,}125$ finger samples.
Both estimators produce qualitatively similar long-tailed distributions on both components.
The corpus-adaptive radial-basis bandwidth resolves to $\hat{\sigma}_W = 24.95$\,mm and $\hat{\sigma}_F = 4.32$\,mm for WiLoR, and to $\hat{\sigma}_W = 28.68$\,mm and $\hat{\sigma}_F = 4.33$\,mm for HaMeR.
The slightly larger $\hat{\sigma}_W$ on HaMeR tracks its slightly higher $80$th-percentile wrist error.

We evaluate four configurations of upstream-estimator handling on the full HOT3D test split (Tab.~\ref{tab:supp_e2_dropin}), reporting the standard 16-joint MANO MPJPE for direct comparison with Tab.~\ref{table:hot3d}.
Row 1 anchors the WiLoR-trained baseline, rows 2 and 3 isolate \del{zero-shot}\rev{uncalibrated} drop-in and bandwidth-recalibrated drop-in with HaMeR observations at inference, and row 4 retrains the entire pipeline (DiT + QN + adaptive $\hat{\sigma}$) end-to-end on HaMeR observations.

\begin{table}[h]
\centering
\caption{\textbf{HaMeR drop-in and full-retrain evaluation on HOT3D.}
Same Ours pipeline architecture, four configurations of the upstream observation source and training scheme: WiLoR-trained baseline, HaMeR \protect\del{zero-shot}\protect\rev{without recalibration} at inference, HaMeR with bandwidth recalibration, and HaMeR end-to-end retraining.
Numbers are reported on the full HOT3D test set ($244$ clips, $n{=}485$ hand-clips) under the 16-joint MANO MPJPE convention of Tab.~\ref{table:hot3d}.}
\label{tab:supp_e2_dropin}
\small
\begin{tabular}{lrrrr}
\toprule
Method & W-MPJPE (mm) & WA-MPJPE (mm) & PA-MPJPE (mm) & AccEr (m/s$^2$) \\
\midrule
Ours (WiLoR) & 57.83 & 19.90 & 3.65 & 3.53 \\
Ours (HaMeR, \del{zero-shot}\rev{uncalib.}) & 104.81 ($+46.98$) & 38.50 & 5.83 & 4.10 \\
Ours (HaMeR, $\hat{\sigma}$-recalib.) & 103.84 ($+46.01$) & 38.49 & 5.81 & 4.09 \\
Ours (HaMeR, full retrain) & 62.16 ($+4.33$) & 21.15 & 3.71 & \textbf{3.49} \\
\bottomrule
\end{tabular}

\end{table}

The \del{HaMeR zero-shot}\rev{uncalibrated HaMeR} drop-in degrades W-MPJPE by approximately $47$\,mm relative to the WiLoR baseline, an expected gap because the deployable pipeline was trained against WiLoR-distributed observations.
Recomputing $\hat{\sigma}$ on HaMeR data changes the downstream W-MPJPE by only $-0.97$\,mm (well within $\pm 1$\,mm), indicating that the bandwidth-adaptation mechanism operates in a flat regime in this range and is robust to estimator-specific bandwidth shifts but does not by itself close the train-test gap.
\rev{Calibration updates no network parameter and does not compensate for a large estimator-induced observation shift.}

Retraining the entire pipeline end-to-end on HaMeR observations closes $91\%$ of the \del{zero-shot}\rev{uncalibrated drop-in} gap, bringing W-MPJPE to $62.16$\,mm (only $+4.33$\,mm above the WiLoR baseline).
The remaining $4.33$\,mm gap reflects a regressor-quality difference rather than a framework limitation: HaMeR's training corpus does not include Aria-style egocentric imagery, while WiLoR's training corpus includes H2O~\cite{kwon2021h2o} which provides comparable egocentric supervision.
PA-MPJPE is essentially unchanged under HaMeR retraining ($+0.06$\,mm, within noise), confirming that the gap is concentrated in wrist absolute positioning rather than finger articulation.
AccEr is slightly better under HaMeR retraining ($-0.04$\,m/s$^2$), suggesting comparable temporal smoothness.

\section[Calibration-Only Transfer]{\rev{Calibration-Only Transfer}}
\label{supp:transfer}

{\revcolor
We evaluate transfer to H2O~\cite{kwon2021h2o} and HOI4D~\cite{liu2022hoi4d} with a quality network retrained after removing both corpora from its pretraining mixture and a HOT3D generative model trained on its predictions.
Both networks stay frozen on the targets, and the only target-domain operation is the estimation of $\hat{\sigma}$ in Eq.~\ref{eq:deploy_sigma} from unlabeled clips, without ground truth or parameter updates.
\par}

\begin{table}[h]
\centering
\revcolor
\caption{\textbf{Calibration-only transfer to H2O and HOI4D.}
$\hat{\sigma}$ is estimated from ten unlabeled target clips, and all methods run on the same clips and hardware.
Per-target best values are in bold.}
\label{tab:supp_transfer}
\small
\resizebox{\linewidth}{!}{%
\begin{tabular}{llccccc}
\toprule
Target & Method & W-MPJPE [mm] ($\downarrow$) & WA-MPJPE [mm] ($\downarrow$) & PA-MPJPE [mm] ($\downarrow$) & AccEr [m/s$^2$] ($\downarrow$) & Time/clip [s] ($\downarrow$) \\
\midrule
\multirow{3}{*}{H2O~\cite{kwon2021h2o}} & Dyn-HaMR~\cite{yu2025dyn} & 60.88 & 52.07 & 19.72 & 3.91 & 323.89 \\
 & HaWoR~\cite{zhang2025hawor} & 97.91 & 60.18 & 12.35 & 9.18 & 31.53 \\
 & Ours & \textbf{56.77} & \textbf{33.94} & \textbf{5.35} & \textbf{1.82} & \textbf{30.62} \\
\midrule
\multirow{3}{*}{HOI4D~\cite{liu2022hoi4d}} & Dyn-HaMR~\cite{yu2025dyn} & \textbf{78.68} & 71.08 & 57.93 & 4.28 & 331.74 \\
 & HaWoR~\cite{zhang2025hawor} & 142.15 & \textbf{41.20} & \textbf{8.13} & 5.43 & 35.08 \\
 & Ours & 133.81 & 77.96 & 8.65 & \textbf{3.49} & \textbf{32.46} \\
\bottomrule
\end{tabular}
}
\end{table}

{\revcolor
On H2O, StableHand is best in every column and runs about $10\times$ faster than Dyn-HaMR, which optimizes each clip.
On HOI4D, articulation and smoothness transfer, with PA-MPJPE close to HaWoR and the lowest AccEr, while W-MPJPE is high for both single-pass camera estimators, ours and HaWoR.
HOI4D head motion is nearly static and leaves little parallax for monocular camera estimation, and the per-clip optimization of Dyn-HaMR attains the lowest W-MPJPE at about $10\times$ our runtime.
\par}

{\revcolor
\noindent\textbf{Calibration-subset size.}
Tab.~\ref{tab:supp_calib_size} varies the number $N$ of unlabeled clips used to estimate $\hat{\sigma}$.
The spread over draws shrinks with $N$, and five clips suffice on H2O, whereas HOI4D needs about thirty.
Without calibration, W-MPJPE is $59.92$\,mm on H2O and $143.39$\,mm on HOI4D, against $53.22$ and $125.35$\,mm with an oracle bandwidth.
\par}

\begin{table}[h]
\centering
\revcolor
\caption{\textbf{Calibration-subset size.}
Mean and standard deviation over five random draws of $N$ unlabeled target clips.}
\label{tab:supp_calib_size}
\small
\resizebox{\linewidth}{!}{%
\begin{tabular}{rcccc}
\toprule
$N$ clips & H2O $\hat{\sigma}_W$ / $\hat{\sigma}_F$ [mm] & H2O W-MPJPE [mm] ($\downarrow$) & HOI4D $\hat{\sigma}_W$ / $\hat{\sigma}_F$ [mm] & HOI4D W-MPJPE [mm] ($\downarrow$) \\
\midrule
1  & $13.5 \pm 5.1$ / $2.42 \pm 0.19$ & $59.60 \pm 1.97$ & $34.6 \pm 11.1$ / $5.24 \pm 2.53$ & $136.68 \pm 13.28$ \\
2  & $16.9 \pm 6.0$ / $2.65 \pm 0.20$ & $56.96 \pm 1.76$ & $43.3 \pm 20.6$ / $3.97 \pm 0.79$ & $140.37 \pm 7.11$ \\
5  & $16.4 \pm 3.3$ / $2.48 \pm 0.05$ & $57.24 \pm 0.91$ & $40.3 \pm 10.8$ / $5.55 \pm 1.65$ & $133.28 \pm 8.67$ \\
10 & $16.6 \pm 1.5$ / $2.54 \pm 0.10$ & $57.13 \pm 0.50$ & $41.6 \pm 9.6$ / $5.30 \pm 1.57$ & $133.64 \pm 7.61$ \\
30 & $13.7 \pm 0.7$ / $2.50 \pm 0.08$ & $57.95 \pm 0.66$ & $48.5 \pm 8.6$ / $5.44 \pm 0.19$ & $132.20 \pm 1.20$ \\
\bottomrule
\end{tabular}
}
\end{table}

\section{Supplemental Ablation Studies}
\label{supp:ablation}

This section reports a complementary ablation analysis to the main-paper studies of Sec.~\ref{subsec:exp_ablation}.
The RBF bandwidth sensitivity in Sec.~\ref{supp:sigma_sweep} sweeps fixed $\sigma$ around our adaptive calibration of Eq.~\ref{eq:deploy_sigma}, and the quality-network ablation in Sec.~\ref{supp:qn_ablation} reports per-corpus Spearman correlation and aggregate metrics across architectural and corpus variants.
\rev{Sec.~\ref{supp:qn_extensions} and Sec.~\ref{supp:granularity} evaluate additional quality-network inputs and finer quality granularity.}

\subsection{RBF Bandwidth Sensitivity}
\label{supp:sigma_sweep}

Tab.~\ref{table:supp_sigma_sweep} sweeps the RBF bandwidth $\sigma$ around the adaptive value $\sigma_\star = p_{80}(e_\star)/\ln 10$ used in the main paper.
Each row retrains the generative model with the labeled $\sigma$ applied to Eqs.~\ref{eq:q_wrist}--\ref{eq:q_finger}, and uses the oracle per-component $\mathbf{q}$ at inference under the same $\sigma$ to isolate the generative-model response to bandwidth choice from quality-network prediction.

\begin{table}[t]
\centering
\caption{\textbf{Sensitivity to the RBF bandwidth $\sigma$ on HOT3D.}
Each row retrains the generative model with the labeled $\sigma$ applied to Eqs.~\ref{eq:q_wrist}--\ref{eq:q_finger}.
Inference uses the oracle per-component $\mathbf{q}$ computed under the same $\sigma$, isolating the generative-model response to bandwidth choice from quality-network prediction.
Per-column \colorbox{bestcolor}{\textbf{best}} and \colorbox{secondcolor}{second best} are color-coded.}
\label{table:supp_sigma_sweep}
\resizebox{0.85\linewidth}{!}{
\setlength{\tabcolsep}{2.5mm}{
\begin{tabular}{lcccc}
\toprule
Setup & PA-MPJPE $\downarrow$ & W-MPJPE $\downarrow$ & WA-MPJPE $\downarrow$ & AccEr $\downarrow$ \\
\midrule
$\sigma$ = adaptive $p_{80}/\ln 10$ (Ours)   & \best{4.02} & \best{57.83} & \best{21.02} & 3.83 \\
$\sigma$ = fixed $30$\,mm                    & \second{4.54} & \second{65.52} & \second{23.36} & \best{3.31} \\
$\sigma$ = fixed $50$\,mm                    & 5.79 & 87.51 & 32.02 & 3.44 \\
$\sigma$ = fixed $80$\,mm                    & 6.67 & 118.70 & 41.56 & \second{3.43} \\
$\sigma$ = fixed $150$\,mm                   & 6.86 & 128.04 & 45.48 & 4.40 \\
\bottomrule
\end{tabular}
}}
\end{table}

W-MPJPE degrades monotonically as $\sigma$ grows.
The adaptive $\sigma_\star$ on HOT3D resolves to $\sigma_W{\approx}26.2$\,mm and $\sigma_F{\approx}6.4$\,mm, close to the fixed $\sigma{=}30$\,mm row but with a per-component split that the fixed sweep cannot reproduce.
Larger $\sigma$ flattens $\exp(-e/\sigma)$ toward $1$ and erodes the contrast between high- and low-quality frames, removing the gradient that the per-channel forward schedule relies on.
AccEr behaves non-monotonically: a moderately wider $\sigma{=}30$\,mm attains the lowest AccEr ($3.31$ vs. $3.83$\,m/s$^2$ for the adaptive $\sigma_\star$) at the cost of a $7.7$\,mm rise in W-MPJPE.
This mirrors the multi-metric trade-off in the ODE-step sweep of Sec.~\ref{supp:runtime}, where joint-position accuracy and trajectory smoothness pull against each other and motivate selecting $\sigma_\star$ on the joint-position metric while reporting both.
Each variant trains and evaluates under the same $\sigma$, matching the deployment scenario but conflating the model's representational adaptation to the bandwidth with the change in $\mathbf{q}$ distribution it induces.

\subsection{Quality Network Ablation}
\label{supp:qn_ablation}

Tab.~\ref{table:ablation}(a) of the main paper isolates the impact of two quality-network design choices on downstream world space hand recovery.
This subsection drills further into the quality network's own predictive fidelity, evaluated by the Spearman rank correlation $\rho(\hat{\mathbf{q}}, \mathbf{q})$ between predicted and ground-truth per-component quality on held-out test splits of multiple egocentric corpora (Tab.~\ref{table:qn_ablation}).

\begin{table}[t]
\centering
\caption{\textbf{Quality-network ablation evaluated by per-corpus Spearman correlation $\rho(\hat{\mathbf{q}}, \mathbf{q})$ and aggregate operational metrics.}
Per-corpus $\rho$ measures rank agreement between predicted and ground-truth per-component quality on each named held-out test split.
$\rho{=}1$ matches the ground-truth ranking and $\rho{=}0$ is random.
Aggregate metrics report q-MAE (mean absolute error of $\hat{\mathbf{q}}$ against the ground-truth $\mathbf{q}$ on the 8-corpus mixed validation), mm-MAE (best $\hat{\mathbf{e}}$ MAE on each ckpt's own training-time validation set, in mm), and gap-closed (the QN-vs-random fraction of the random-to-oracle quality-rejection AUC gap on the mixed validation).
(a)~Ablates the input cues, temporal architecture, and training loss, holding the pretraining corpus fixed.
The Ours and \emph{w/o Wasserstein loss} rows are evaluated under the v26 protocol while the architectural variants retain their original v22 measurements (cross-protocol comparison on a few cells is therefore loose).
(b)~Ablates the pretraining corpus, holding the architecture and training loss fixed.
The mm-MAE column uses different validation sets for the two rows (HOT3D-only val for the single-corpus row, mixed val for the multi-corpus row), so cross-row comparison on this column is loose.
Per-column \colorbox{bestcolor}{\textbf{best}} and \colorbox{secondcolor}{second best} within each block are color-coded.}
\label{table:qn_ablation}
\resizebox{\linewidth}{!}{
\setlength{\tabcolsep}{1.4mm}{
\begin{tabular}{lcccccccccc}
\toprule
& \multicolumn{6}{c}{Per-corpus Spearman $\rho$ $\uparrow$} & \multicolumn{3}{c}{Aggregate} \\
\cmidrule(lr){2-7} \cmidrule(lr){8-10}
Setup & HOT3D & H2O & HOI4D & Re:InterHand & ARCTIC & Avg & q-MAE $\downarrow$ & mm-MAE (mm) $\downarrow$ & gap-closed $\uparrow$ \\
\midrule
\multicolumn{10}{l}{\emph{(a) Quality-network architecture and training loss}} \\
\textbf{Full QN (Ours)}                                & \second{0.846} & \best{0.906} & \best{0.859} & \second{0.912} & \best{0.775} & \best{0.862} & \second{0.096} & \second{12.55} & \best{+98.5\%} \\
+ visual feature $\mathbf{f}^h_t$ (no exclusion)       & 0.775 & 0.855 & 0.814 & 0.883 & 0.649 & 0.825 & 0.120 & 24.07 & +86.9\% \\
w/o camera-pose input $\mathbf{g}_t$                   & 0.797 & 0.857 & 0.801 & 0.880 & 0.667 & 0.836 & 0.116 & 27.90 & +88.5\% \\
w/o detection flag $\delta^h_t$                        & \best{0.847} & \second{0.887} & 0.834 & \best{0.917} & 0.613 & \second{0.859} & 0.104 & 18.64 & +89.6\% \\
per-frame MLP (no temporal self-attention)             & 0.819 & 0.857 & 0.830 & 0.892 & 0.697 & 0.838 & \best{0.078} & 22.10 & +89.6\% \\
w/o Wasserstein loss (MSE-only training)               & 0.831 & 0.880 & \second{0.840} & 0.907 & \second{0.754} & 0.844 & \second{0.096} & \best{11.04} & \second{+91.7\%} \\
\midrule
\multicolumn{10}{l}{\emph{(b) Pretraining corpus}} \\
QN pretrained on HOT3D only                            & \best{0.850} & 0.707 & 0.766 & 0.704 & 0.623 & 0.757 & 0.162 & \best{9.48} & +83.9\% \\
\textbf{QN pretrained on full multi-corpus mixture (Ours)} & \second{0.846} & \best{0.906} & \best{0.859} & \best{0.912} & \best{0.775} & \best{0.862} & \best{0.096} & 12.55 & \best{+98.5\%} \\
\bottomrule
\end{tabular}
}}
\end{table}

\noindent\textbf{Architecture and training loss (Block (a)).}
Each row of Tab.~\ref{table:qn_ablation}(a) ablates one design choice of the quality network.
Adding the frozen WiLoR visual feature $\mathbf{f}^h_t$ back to the per-hand input token degrades cross-corpus average $\rho$ by $0.037$ and lifts the validation $\hat{\mathbf{e}}$ MAE from $12.55$\,mm to $24.07$\,mm.
Even under the dataset-invariant log-error target, the 1280-dimensional ViT feature inflates input redundancy and dilutes the optimization signal rather than supplying a complementary cue.
Removing the camera-pose input $\mathbf{g}_t$ costs $0.026$ average $\rho$, with consistent drops of $0.05$ to $0.10$ on every in-train corpus, confirming that head-motion context contributes broadly to per-component error prediction across egocentric streams.
Removing the binary hand-confidence flag $\delta^h_t$ costs only $0.003$ average $\rho$, indicating that the flag is a near-redundant input recoverable from the proposal vector $\bar{\mathbf{y}}^h_t$ and its temporal context.
Replacing the temporal self-attention encoder with a per-frame MLP raises the validation $\hat{\mathbf{e}}$ MAE from $12.55$\,mm to $22.10$\,mm and trims average $\rho$ by $0.024$, leaving rank correlation almost unchanged but losing the magnitude-refinement role of cross-frame attention.
The per-frame MLP variant retains a competitive q-MAE ($0.078$ vs.\ $0.096$), suggesting that the leaner predictor overfits less to in-train corpora at the cost of magnitude refinement.
Removing the Wasserstein loss of Eq.~\ref{eq:qn_loss} (MSE-only training) lowers the validation $\hat{\mathbf{e}}$ MAE marginally to $11.04$\,mm but collapses the operational gap-closed metric from $+98.5\%$ to $+91.7\%$ and the cross-corpus average $\rho$ from $0.862$ to $0.844$.
This isolates the Wasserstein term's role: it trades a small per-sample magnitude regression for a large gain in distribution alignment, directly driving the deployment-time match between predicted and oracle $\sigma$.
Across the architectural ablations, every input or encoder choice contributes between $0.003$ and $0.037$ average $\rho$, all dominated by the $0.105$ swing between the multi-corpus and single-corpus pretraining of Block (b).

\noindent\textbf{Pretraining corpus (Block (b)).}
A single-corpus quality network pretrained on HOT3D alone marginally beats the multi-corpus model on its own training distribution ($\rho{=}0.850$ vs.\ $0.846$ on HOT3D).
On every other corpus the single-corpus model degrades, with $\rho \in [0.704, 0.766]$ on H2O, HOI4D, and Re:InterHand and $\rho{=}0.623$ on the \del{zero-shot}\rev{held-out} ARCTIC test split.
The multi-corpus model lifts these out-of-distribution cells to $\rho \geq 0.859$ on H2O, HOI4D, and Re:InterHand and to $0.775$ on ARCTIC, raising the cross-corpus average from $0.757$ to $0.862$.
Calibration degrades more sharply than rank: the single-corpus q-MAE averages $0.162$ against the multi-corpus model's $0.096$, and the operational gap-closed metric drops from $+98.5\%$ to $+83.9\%$.
Broad pretraining is essential for the quality network to produce $\hat{\mathbf{q}}$ that is both well-ranked and well-calibrated on out-of-distribution egocentric corpora at deployment time.
\rev{Sec.~\ref{supp:transfer} evaluates the corresponding full-system transfer.}

\subsection[Quality-Network Input Extensions]{\rev{Quality-Network Input Extensions}}
\label{supp:qn_extensions}

{\revcolor
The quality network of the main paper receives the hand observation, the detection flag, and the camera pose.
We test two additional inputs that avoid the dataset shortcut of the frozen WiLoR feature in Tab.~\ref{table:qn_ablation}(a).
The first adds six texture-free SAM3 mask statistics per hand and frame (visible area, mask confidence, hand-object overlap, box IoU, area change, and a missing-mask flag).
With HOI4D held out, it raises $\rho$ from $0.382$ to $0.553$ and lowers the error MAE from $48.4$ to $39.5$\,mm on HOI4D, at $2.5\times$ the runtime (Tab.~\ref{tab:supp_mask}).
The second adds a hand-scene depth residual, the median absolute disagreement between the estimated hand-joint depth and the Depth-Anything-V3 depth at the hand, and is compared with the global scene token of the same model.
With H2O held out and a generative model trained for each variant, the scene token degrades W-MPJPE, whereas the depth residual lowers it by $1.53$\,mm on HOT3D and $5.66$\,mm on H2O (Tab.~\ref{tab:supp_scene}).
We keep the lower-cost pose-only quality network as the default and report both inputs as optional extensions.
\par}

\begin{table}[h]
\centering
\revcolor
\caption{\textbf{Mask statistics as quality-network input.}
Wrist channel with HOI4D held out.
$\rho$ is the within-clip Spearman correlation, and AUROC measures the detection of high-error frames.}
\label{tab:supp_mask}
\small
\resizebox{\linewidth}{!}{%
\begin{tabular}{lcccc}
\toprule
QN input & HOT3D $\rho$ / AUROC ($\uparrow$) & HOI4D $\rho$ / AUROC ($\uparrow$) & HOI4D MAE [mm] ($\downarrow$) & Time/clip [s] ($\downarrow$) \\
\midrule
Pose-only (Ours) & \textbf{0.439} / \textbf{0.799} & 0.382 / 0.516 & 48.4 & \textbf{27.29} \\
+ mask statistics & 0.414 / 0.797 & \textbf{0.553} / \textbf{0.658} & \textbf{39.5} & 67.83 \\
\bottomrule
\end{tabular}
}
\end{table}

\begin{table}[h]
\centering
\revcolor
\caption{\textbf{Scene-geometry inputs to the quality network.}
Wrist channel with HOT3D in domain and H2O held out.
The last two columns report the downstream W-MPJPE.}
\label{tab:supp_scene}
\small
\resizebox{\linewidth}{!}{%
\begin{tabular}{lccccc}
\toprule
QN input & Val MAE [mm] ($\downarrow$) & HOT3D $\rho$ / AUROC ($\uparrow$) & H2O $\rho$ / AUROC ($\uparrow$) & HOT3D W-MPJPE [mm] ($\downarrow$) & H2O W-MPJPE [mm] ($\downarrow$) \\
\midrule
Pose-only (Ours) & 10.98 & \textbf{0.439} / \textbf{0.799} & 0.361 / 0.496 & 57.83 & 56.77 \\
+ scene token & 11.31 & 0.416 / 0.780 & \textbf{0.533} / \textbf{0.547} & 64.43 & 60.13 \\
+ depth residual & \textbf{10.94} & 0.409 / 0.792 & 0.520 / 0.510 & \textbf{56.30} & \textbf{51.11} \\
\bottomrule
\end{tabular}
}
\end{table}

\subsection[Quality Granularity]{\rev{Quality Granularity}}
\label{supp:granularity}

{\revcolor
Tab.~\ref{tab:supp_granularity} evaluates finer quality channels on the $155$ ARCTIC s05 clips.
Per-finger channels give modest gains on all five metrics, including $3.96$\,mm lower W-MPJPE, and per-joint channels add no consistent gain beyond them.
Only $8.2\%$ of the within-frame joint-error variance lies across finger identity, against $83.4\%$ across kinematic depth, and we keep the compact wrist and finger channels of the main paper.
\par}

\begin{table}[h]
\centering
\revcolor
\caption{\textbf{Quality granularity on ARCTIC.}
MPJPE and MRRPE are in mm and AccEr in m/s$^2$.}
\label{tab:supp_granularity}
\small
\resizebox{\linewidth}{!}{%
\begin{tabular}{lcccccc}
\toprule
Granularity & Channels per hand & PA-MPJPE ($\downarrow$) & W-MPJPE ($\downarrow$) & WA-MPJPE ($\downarrow$) & AccEr ($\downarrow$) & MRRPE ($\downarrow$) \\
\midrule
Wrist + fingers (Ours) & 2 & 8.07 & 124.59 & 32.66 & 4.67 & 93.71 \\
Per-finger & 6 & \textbf{7.87} & \textbf{120.63} & 31.79 & \textbf{4.31} & \textbf{90.20} \\
Per-joint & 16 & 8.62 & 121.85 & \textbf{31.61} & 5.91 & 92.36 \\
\bottomrule
\end{tabular}
}
\end{table}

\section{Inference Efficiency}
\label{supp:runtime}

Tab.~\ref{table:runtime} reports wall-clock time per $150$-frame clip on a single H100 GPU.
\del{Optimization-based baselines (HaWoR, Dyn-HaMR) require}\rev{The optimization-based Dyn-HaMR requires} per-clip SLAM and iterative refinement, while StableHand runs a single quality-network forward pass followed by a $20$-step Euler ODE solve with no per-clip optimization loop.
The runtime gap between StableHand and SLAM-based baselines is driven primarily by the choice of geometry foundation model (DA3 forward pass vs \rev{DROID-SLAM} optimization) rather than by the quality-aware flow-matching pathway itself, and we report it here for completeness rather than as a contribution of this work.

\begin{table}[t]
\centering
\caption{\textbf{Inference efficiency on HOT3D} ($150$-frame clips, single H100).
We report wall-clock time per clip and peak GPU memory.
\protect\del{Optimization-based baselines are}\protect\rev{The optimization-based Dyn-HaMR is} an order of magnitude slower than feed-forward generative models.
StableHand stays in the feed-forward envelope despite solving a $20$-step ODE.
Per-column \colorbox{bestcolor}{\textbf{best}} and \colorbox{secondcolor}{second best} are color-coded.}
\label{table:runtime}
\resizebox{0.65\linewidth}{!}{
\setlength{\tabcolsep}{2.5mm}{
\begin{tabular}{lcc}
\toprule
Method & Time / clip (s) $\downarrow$ & Peak GPU mem (GB) $\downarrow$ \\
\midrule
HaMeR-SLAM~\cite{pavlakos2024reconstructing}  & 125.91 & 12.50 \\
WiLoR-SLAM~\cite{potamias2025wilor}           & 30.12  & 12.50 \\
\midrule
Dyn-HaMR~\cite{yu2025dyn}                     & 347.84 & \best{9.82} \\
HaWoR~\cite{zhang2025hawor}                   & \second{28.45}  & 14.56 \\
\textbf{StableHand (Ours)}                    & \best{27.29} & \second{10.32} \\
\bottomrule
\end{tabular}
}}
\end{table}

\noindent\textbf{Per-stage breakdown.}
Tab.~\ref{tab:supp_e9_breakdown} dissects the StableHand wall-clock cost into its four stages.
The pipeline cost is dominated by the geometry foundation model (Depth-Anything-V3, $17.09$\,s per clip) and the per-frame WiLoR estimator ($9.96$\,s per clip), while the quality network ($<\!0.01$\,s) and the $20$-step Euler ODE solve through the DiT denoiser ($0.23$\,s) together account for less than $1\%$ of the total wall-clock time.
The quality-aware flow-matching pathway is essentially free relative to the upstream visual stack.

\begin{table}[h]
\centering
\caption{\textbf{Per-stage inference-time breakdown on a single H100 GPU.}
Average wall-clock time per $T{=}150$ clip, averaged over $10$ HOT3D clips with batch size $1$.
Per-stage values are proportionally aligned to the wall-clock total of Tab.~\ref{table:runtime}.}
\label{tab:supp_e9_breakdown}
\small
\begin{tabular}{lrr}
\toprule
Stage & Time per clip (s) & Time per frame (ms) \\
\midrule
Depth-Anything-V3~\cite{lin2025depth} (geometry, $1$ call) & 17.09 & 113.93 \\
WiLoR~\cite{potamias2025wilor} estimator ($\times 150$ frames) & 9.96 & 66.40 \\
DiT $\times$ $20$ ODE steps & 0.23 & 1.51 \\
Quality network ($1$ forward over $T$) & $<\!0.01$ & $<\!0.10$ \\
\midrule
\textbf{Total (full pipeline)} & \textbf{27.29} & \textbf{181.93} \\
\bottomrule
\end{tabular}

\end{table}

\begin{figure}[h]
  \centering
  \includegraphics[width=0.85\linewidth]{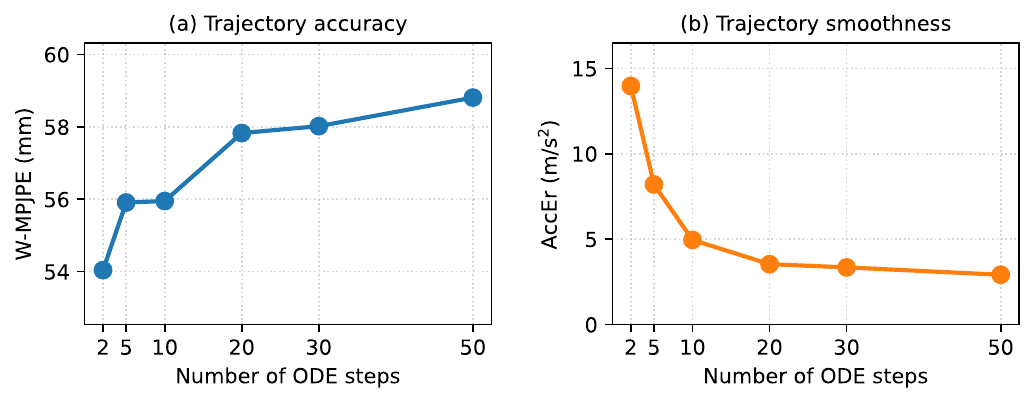}
  \caption{\textbf{ODE-step sweep on HOT3D test, multi-metric view.}
  $n_\mathrm{steps}\in\{2,5,10,20,30,50\}$ evaluated on (a) trajectory accuracy (W-MPJPE) and (b) trajectory smoothness (AccEr), with the $n{=}20$ row anchored to the main paper Tab.~\ref{table:hot3d}.
  Lower step counts attain marginally lower W-MPJPE (under $9\%$ range) but degrade AccEr by nearly $4\times$ at $n{=}2$, motivating $n{=}20$ as the deployable default where AccEr is largely converged.}
  \label{fig:supp_e9_ode_sweep}
\end{figure}

\noindent\textbf{ODE-step sweep.}
Fig.~\ref{fig:supp_e9_ode_sweep} sweeps the Euler ODE step count $n_\mathrm{steps}\in\{2,5,10,20,30,50\}$ on HOT3D test, evaluated jointly on trajectory accuracy (W-MPJPE) and trajectory smoothness (AccEr).
The two metrics behave oppositely with $n_\mathrm{steps}$: W-MPJPE varies by under $9\%$ across the entire sweep ($54.04$\,mm at $n{=}2$ to $58.81$\,mm at $n{=}50$), while AccEr drops nearly $4\times$ from $13.97$ at $n{=}2$ to $3.53$\,m/s$^2$ at $n{=}20$ as the dense Euler integration replaces stair-step trajectory reconstruction with a smooth one.
Most of the AccEr improvement occurs by $n{=}20$, with diminishing returns thereafter ($3.34$ and $2.91$\,m/s$^2$ at $n{=}30$ and $n{=}50$ at $1.5\times$ and $2.5\times$ the DiT compute respectively).
We adopt $n{=}20$ as the deployable default since temporal smoothness is largely converged at this setting and the marginal gain from higher step counts does not justify the extra cost, while the remaining real-time bottleneck lies in the upstream pose estimator and geometry foundation model rather than in the quality-aware flow-matching pathway proposed in this work.

\section{Synthetic Perturbation Taxonomy}
\label{supp:augmentation}

This section formalizes the quality-consistent observation augmentation introduced in Sec.~\ref{sec:method_qsignal}.
We define the augmentation as a stochastic operator on $(\bar{\mathbf{y}}^h_t,\mathbf{q}^h_t)$ pairs, derive the joint update rule that preserves the calibration of Eq.~\ref{eq:deploy_sigma}, and describe each of the five perturbation types together with the upstream-stream failure mode it imitates.

\noindent\textbf{Augmentation operator.}
Let $\mathcal{A}$ denote the augmentation operator, applied independently per hand $h\!\in\!\{L,R\}$ at every training step.
With probability $\tfrac{1}{2}$, $\mathcal{A}$ leaves the input unchanged.
Otherwise, $\mathcal{A}$ samples a perturbation type uniformly from a five-element pool and applies it to a designated channel subset of the per-hand observation $\bar{\mathbf{y}}^h_t\!\in\!\mathbb{R}^{54}$, partitioned as $\bar{\mathbf{y}}^h_t\!=\![\bar{\boldsymbol{\theta}}^h_{\!\mathrm{wrist}};\,\bar{\boldsymbol{\theta}}^h_{\!\mathrm{fingers}};\,\bar{\mathbf{p}}^h]$ with $\bar{\boldsymbol{\theta}}^h_{\!\mathrm{wrist}}\!\in\!\mathbb{R}^{6}$ (rot6d), $\bar{\boldsymbol{\theta}}^h_{\!\mathrm{fingers}}\!\in\!\mathbb{R}^{45}$ (finger axis-angles), and $\bar{\mathbf{p}}^h\!\in\!\mathbb{R}^{3}$ (wrist translation).
The visual feature $\mathbf{f}^h_t$, camera pose $\mathbf{g}_t$, and scene-geometry token $\mathbf{s}_t$ are never perturbed.

\noindent\textbf{Joint quality update.}
Continuous perturbations attenuate the affected quality channel multiplicatively,
\begin{equation}
q^{h\prime}_\star \;=\; q^h_\star\cdot \exp\!\left(-\Delta e_\star / \sigma_\star\right),
\qquad \star\!\in\!\{W,F\},
\label{eq:supp_q_update_cont}
\end{equation}
where $\sigma_W,\sigma_F$ are the per-corpus bandwidths of Eq.~\ref{eq:deploy_sigma} and $\Delta e_\star\!\geq\!0$ is the perturbation-induced increase in the corresponding component error of Eqs.~\ref{eq:q_wrist}--\ref{eq:q_finger}.
Eq.~\ref{eq:supp_q_update_cont} follows from the RBF form $q\!=\!\exp(-e/\sigma)$: substituting the post-perturbation error $e_\star\!+\!\Delta e_\star$ factors $q^{h\prime}_\star$ into $q^h_\star\!\cdot\!\exp(-\Delta e_\star/\sigma_\star)$.
Discrete perturbations instead clamp the affected quality channels to zero,
\begin{equation}
q^{h\prime}_\star \;=\; 0,
\label{eq:supp_q_update_disc}
\end{equation}
mirroring the inference-time treatment of frames whose hand-confidence flag $\delta^h_t\!=\!0$ or whose pose collapses to a degenerate configuration.

\noindent\textbf{Gaussian noise.}
We sample a noise scale $\sigma_g\!\sim\!\mathrm{Unif}\{10^{-2},2\!\times\!10^{-2},5\!\times\!10^{-2}\}$ and add zero-mean isotropic Gaussian noise of standard deviation $\sigma_g$ to all $54$ channels of $\bar{\mathbf{y}}^h_t$.
Both wrist and finger errors increase, attenuating $q^h_W$ and $q^h_F$ via Eq.~\ref{eq:supp_q_update_cont}.
This type imitates the residual prediction noise that MANO regression heads incur on textured backgrounds.

\noindent\textbf{Depth scaling.}
We sample a scale factor $s\!\sim\!\mathrm{Unif}(0.5,2.0)$ and rescale the wrist translation as $\bar{\mathbf{p}}^{h\prime}\!=\!s\,\bar{\mathbf{p}}^h$, leaving wrist rotation and finger articulation unchanged.
Only the wrist error increases, attenuating $q^h_W$ via Eq.~\ref{eq:supp_q_update_cont}.
This type imitates the metric-depth ambiguity intrinsic to monocular geometry estimators on egocentric video.
On our deployable pipeline this manifests as DA3~\cite{lin2025depth} drift on novel scenes that shifts the wrist by hundreds of millimeters.

\noindent\textbf{Wrist-rotation jitter.}
We add zero-mean Gaussian noise of standard deviation $\sigma_R\!=\!0.05$\,rad to the rot6d representation of the wrist rotation $\bar{\boldsymbol{\theta}}^h_{\!\mathrm{wrist}}$, leaving wrist translation and finger articulation unchanged.
Since the wrist error of Eq.~\ref{eq:q_wrist} is defined on translation alone, we attenuate $q^h_W$ via the angular variant $q^{h\prime}_W\!=\!q^h_W\cdot\exp(-\theta_R/\sigma_R)$, where $\theta_R$ is the geodesic distance between the perturbed and unperturbed wrist rotations.
This type imitates tracker confusion under fast head rotation.

\noindent\textbf{Finger lock-in.}
We sample a contiguous window of $L\!\sim\!\mathrm{Unif}(5,30)$ frames and clamp the $45$ finger axis-angle channels of $\bar{\boldsymbol{\theta}}^h_{\!\mathrm{fingers}}$ to zero across the window.
The finger quality channel is clamped to $q^{h\prime}_F\!=\!0$ via Eq.~\ref{eq:supp_q_update_disc}.
This type imitates short-window articulation failures during occluded grasps.

\noindent\textbf{Per-hand dropout.}
We sample a contiguous window of $L\!\sim\!\mathrm{Unif}(5,30)$ frames, zero all $54$ channels of $\bar{\mathbf{y}}^h_t$, and reset the hand-confidence flag to $\delta^h_t\!=\!0$ across the window.
Both quality channels are clamped to $q^{h\prime}_W\!=\!q^{h\prime}_F\!=\!0$ via Eq.~\ref{eq:supp_q_update_disc}.
This type imitates extended out-of-frame events when one hand leaves the egocentric camera frustum during bimanual manipulation.

\noindent\textbf{Effect on the training distribution.}
Without augmentation, the calibration of Eq.~\ref{eq:deploy_sigma} places only $\sim\!20\%$ of training frames in the low-quality region $q\!<\!0.1$ on each $q$ component, by construction.
Under the augmentation operator $\mathcal{A}$, this fraction rises to over $50\%$ on each of $q^h_W$ and $q^h_F$.
The per-channel forward process of Sec.~\ref{sec:method_fm} therefore receives substantially broader exposure to the low-quality regime under augmentation.
The increase requires no additional annotated frames: every perturbed sample inherits its ground-truth motion target from the original corpus.

\section{Training Dataset Details}
\label{supp:datasets}

The quality network is pretrained on eight egocentric bimanual hand-pose corpora \rev{totaling} 10M frames, split into two tiers based on annotation type.
\textbf{Tier~1} (native MANO ground truth) comprises five datasets, on which the quality label is computed directly from MANO joint-regression error.
\textbf{Tier~2} (3D joint annotations only) comprises three datasets, on which the quality label is computed from 3D joint MPJPE without any MANO fitting.
All eight corpora are restricted to egocentric viewpoints: for ARCTIC and Ego-Exo4D we use the egocentric stream only, and for Re:InterHand we use the released egocentric camera split.
Tab.~\ref{tab:train_datasets} summarizes scale, viewpoint, and annotation type for each corpus.

\begin{table}[h]
\centering
\caption{\textbf{Training corpora for quality-network pretraining.} All eight datasets are restricted to egocentric viewpoints. Tier~1 datasets provide native MANO parameters. Tier~2 datasets provide 3D joint positions only and supervise the quality label through joint MPJPE without MANO fitting.}
\label{tab:train_datasets}
\small
\begin{tabular}{llrrll}
\toprule
Tier & Dataset & Frames & Subjects & View & GT type \\
\midrule
\multirow{5}{*}{1}
& HOT3D~\cite{banerjee2024introducing}      & 1.5M  & 19  & ego          & MANO \\
& ARCTIC~\cite{fan2023arctic}               & 0.24M & 10  & ego (stream) & MANO \\
& Re:InterHand~\cite{moon2023dataset}       & 0.15M & 26  & ego (synth.) & MANO \\
& H2O~\cite{kwon2021h2o}                    & 2.0M  & 4   & ego          & MANO \\
& HOI4D~\cite{liu2022hoi4d}                 & 0.7M  & 9   & ego          & MANO \\
\midrule
\multirow{3}{*}{2}
& EgoDex~\cite{hoque2025egodex}             & 2.3M  & --   & ego          & 25 joints \\
& EgoVerse~\cite{punamiya2026egoverse}      & 1.5M  & 2087 & ego          & 21 joints \\
& Ego-Exo4D~\cite{grauman2024ego}           & 1.6M  & 740  & ego (stream) & 21 joints \\
\midrule
& \textbf{Total}                            & \textbf{10M} & & & \\
\bottomrule
\end{tabular}
\end{table}

\noindent\textbf{Tier~1 datasets.}

\emph{HOT3D}~\cite{banerjee2024introducing} is the primary benchmark: egocentric clips captured by Project Aria and Quest~3 with MANO annotations obtained from optical-marker motion capture.
The dataset exhibits frequent hand out-of-view events and hand-object occlusion under dynamic egocentric camera motion.

\emph{ARCTIC}~\cite{fan2023arctic} captures bimanual manipulation of articulated objects (scissors, laptops) with one egocentric and eight allocentric cameras.
We use the egocentric stream only.
Both hands are frequently mutually occluded during dexterous manipulation, producing large per-hand quality asymmetries.
We adopt the official ARCTIC P2 protocol, which partitions the ten recording subjects (s01--s10) into eight training subjects (s01, s02, s04, s06, s07, s08, s09, s10), one validation subject (s05), and one test subject (s03) whose ground truth is withheld for the private leaderboard.
Because the s03 ground truth is not publicly available, we report all ARCTIC numbers on the validation subject s05 (34 sequences sliced into $155$ clips of $T{=}150$ frames at $30$\,fps).
The same s05 is held out from the multi-corpus quality-network pretraining.

\emph{Re:InterHand}~\cite{moon2023dataset} provides photorealistically re-rendered InterHand2.6M~\cite{moon2020interhand2} two-hand interaction motions under randomized illumination and camera placement.
We take only the released egocentric camera split (${\sim}147$K frames).
Native MANO parameters are inherited from the source motion capture, and the corpus supplies dense bimanual hand-hand interaction supervision under ego viewpoints, a regime under-represented in the rest of the mixture.

\emph{H2O}~\cite{kwon2021h2o} captures egocentric two-hand object manipulation across 4 subjects, 8 actions, and 36 objects with native MANO annotations and per-frame object 6DoF poses.

\emph{HOI4D}~\cite{liu2022hoi4d} captures 4000 egocentric sequences of category-level human-object interaction across 9 indoor scenes and 16 object categories with a head-mounted Kinect Azure, providing native MANO parameters and 3D joint positions under natural ego head motion.
HOI4D contains a strong right-hand bias (only ${\sim}10\%$ of valid hand-frames are left-hand), which primarily contributes right-hand egocentric supervision to the quality network.

\noindent\textbf{Tier~2 datasets.}

\emph{EgoDex}~\cite{hoque2025egodex} provides 829 hours of egocentric video from Apple Vision Pro with high-precision 25-joint hand tracking across 194 tabletop tasks.

\emph{EgoVerse}~\cite{punamiya2026egoverse} aggregates 1,362 hours of egocentric demonstrations from 2,087 participants across 240 scenes, offering the largest demographic and environmental diversity.

\emph{Ego-Exo4D}~\cite{grauman2024ego} provides 21M automatic 3D hand-joint annotations from synchronized ego and exo cameras across 13 cities.
We use the egocentric stream only.

Tier~2 datasets are used for quality-network pretraining only: the quality label $q{=}\exp({-}e/\sigma)$ is computed from the 3D joint MPJPE between the visual-encoder prediction and the dataset ground truth, requiring no MANO parameter fitting.

\section{Evaluation Metrics}
\label{supp:metrics}

Let $\hat{\mathbf{J}}\in\mathbb{R}^{T\times K\times 3}$ and $\mathbf{J}\in\mathbb{R}^{T\times K\times 3}$ denote the predicted and ground-truth joint positions of a single hand over $T$ frames with $K$ joints.
\rev{We use the $K{=}16$ MANO joints in a gravity-aligned world frame whose origin is the first-frame camera center.}
All metrics are computed per hand on frames with valid ground-truth annotations, then averaged across hands and clips.
\rev{W-MPJPE fits no alignment, and WA-MPJPE subtracts the per-frame wrist joint $k{=}0$,}
\begin{equation}
\revcolor
\begin{aligned}
\mathrm{W\text{-}MPJPE} &= \frac{1}{TK}\sum_{t,k}\|\hat{\mathbf{J}}_{t,k}-\mathbf{J}_{t,k}\|_2,\\
\mathrm{WA\text{-}MPJPE} &= \frac{1}{TK}\sum_{t,k}\bigl\|(\hat{\mathbf{J}}_{t,k}-\hat{\mathbf{J}}_{t,0})-(\mathbf{J}_{t,k}-\mathbf{J}_{t,0})\bigr\|_2.
\end{aligned}
\end{equation}
\rev{W-MPJPE thus retains the accumulated trajectory drift, and WA-MPJPE removes the wrist translation but keeps the global hand orientation.}
PA-MPJPE applies a per-frame Procrustes alignment (rotation, translation, and uniform scale) before measuring joint error, isolating articulation quality.
AccEr is the mean per-joint error of the acceleration $f^2(\mathbf{J}_{t+1}-2\mathbf{J}_{t}+\mathbf{J}_{t-1})$ at $f{=}30$\,fps\rev{, over frame triples with valid ground truth}.
MRRPE~\cite{fan2023arctic,moon2020interhand2} is the error of the relative offset between the two wrists\rev{, averaged over frames where both hands are valid and then over clips}, and is reported on ARCTIC only.

\section{Additional Qualitative Results}
\label{supp:more_qualitative}

\begin{figure*}[!t]
  \centering
  \includegraphics[width=\linewidth]{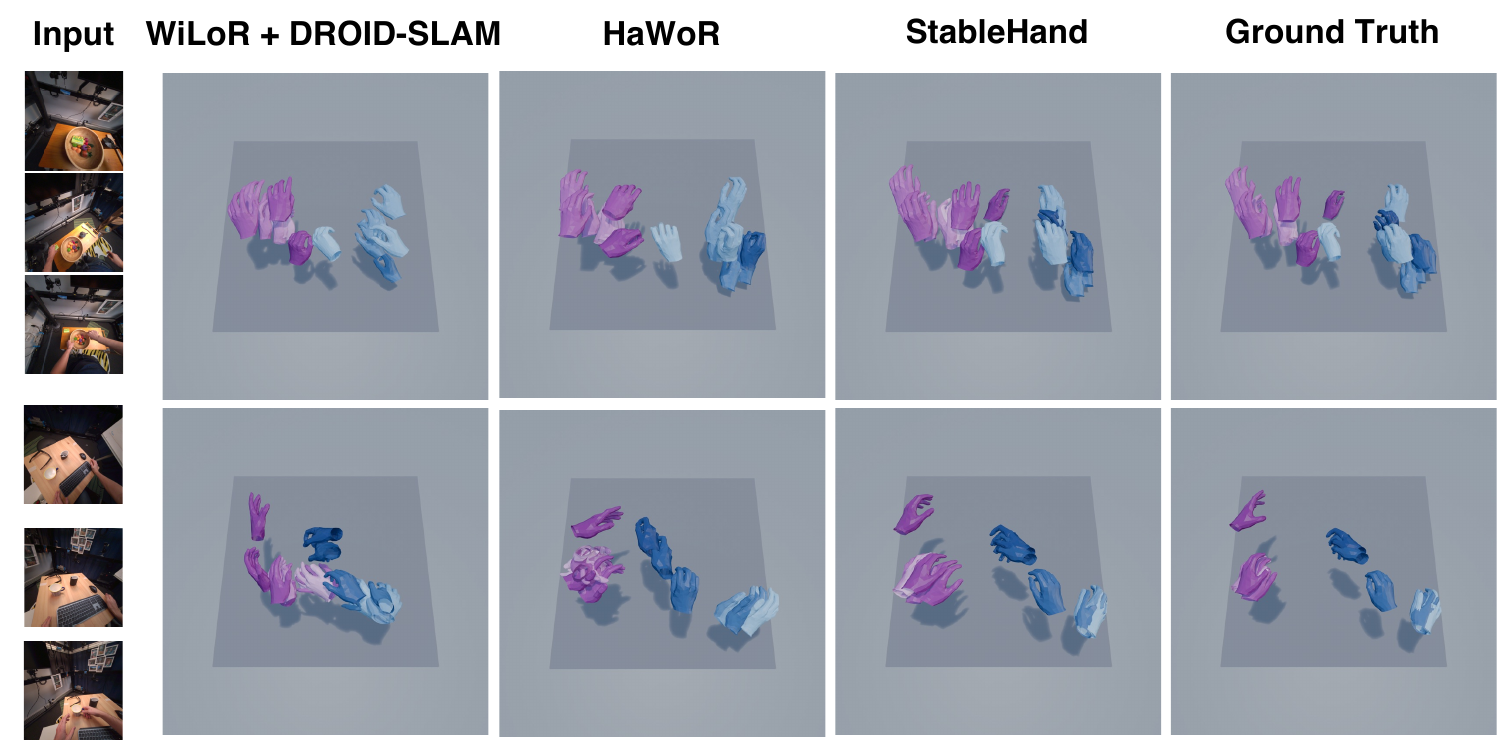}
  \caption{\textbf{Additional qualitative results on HOT3D~\cite{banerjee2024introducing}.}
  Two further HOT3D clips with long missing-hand spans, comparing WiLoR~\cite{potamias2025wilor}+DROID-SLAM~\cite{teed2021droid}, HaWoR~\cite{zhang2025hawor}, our StableHand, and the Ground Truth.
  Layout, color, and temporal-shading conventions follow Fig.~\ref{fig:qualitative}.}
  \label{fig:supp_more_hot3d}
\end{figure*}

\begin{figure*}[!t]
  \centering
  \includegraphics[width=\linewidth]{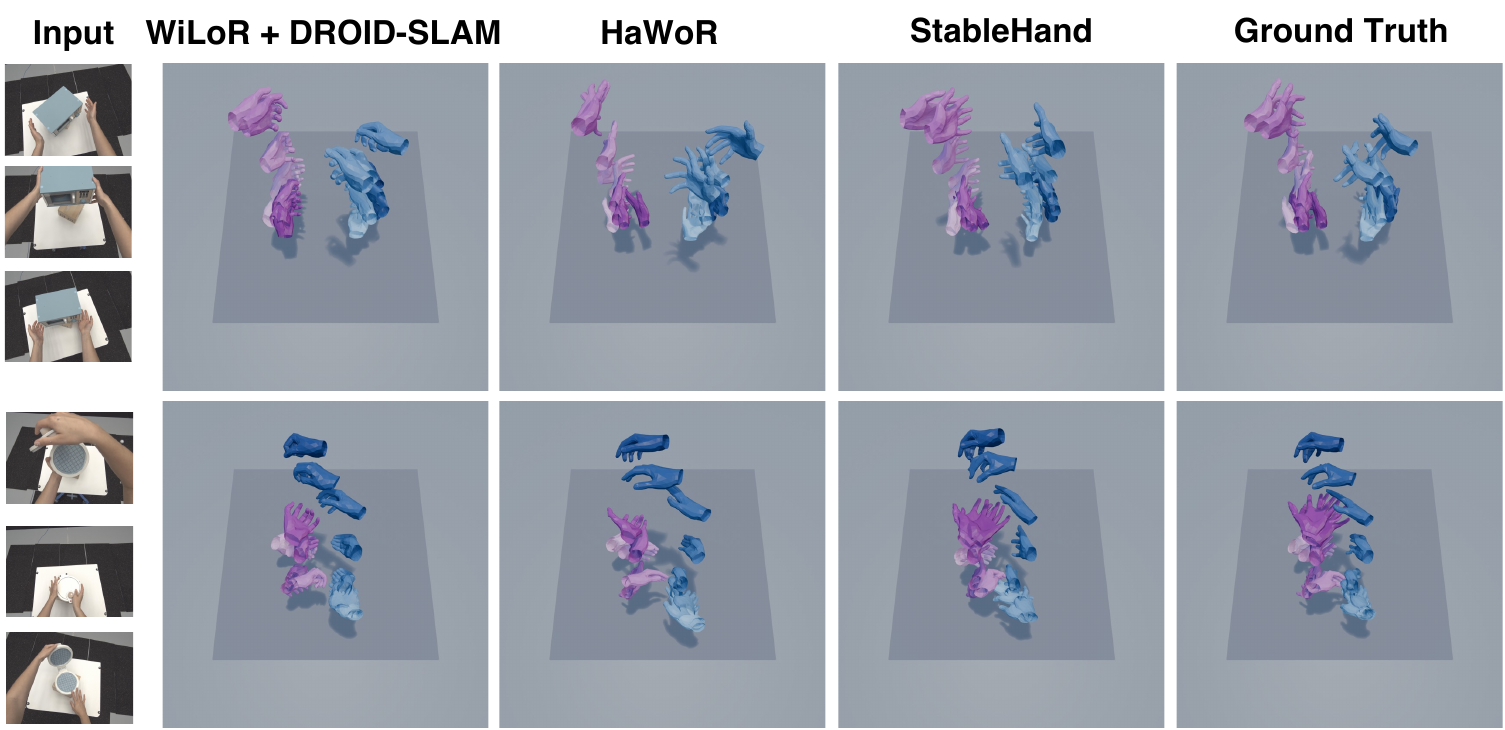}
  \caption{\textbf{Additional qualitative results on ARCTIC~\cite{fan2023arctic}.}
  Two further ARCTIC clips with persistent hand-object occlusion, comparing WiLoR~\cite{potamias2025wilor}+DROID-SLAM~\cite{teed2021droid}, HaWoR~\cite{zhang2025hawor}, our StableHand, and the Ground Truth.
  Layout, color, and temporal-shading conventions follow Fig.~\ref{fig:qualitative}.}
  \label{fig:supp_more_arctic}
\end{figure*}

\begin{figure*}[!t]
  \centering
  \includegraphics[width=\linewidth]{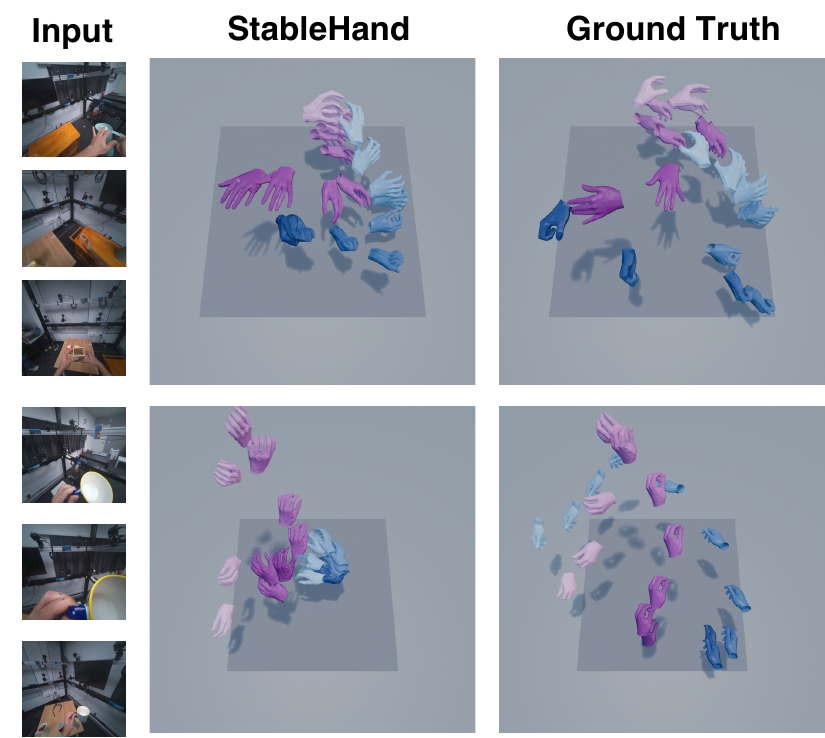}
  \caption{\textbf{Failure cases of StableHand on HOT3D~\cite{banerjee2024introducing}.}
  Each row shows three input frames (left) together with our prediction and the Ground Truth, with color and temporal-shading conventions following Fig.~\ref{fig:qualitative}.
  Top: the left hand leaves the egocentric view for an extended span, leaving its channel unobserved for tens of consecutive frames.
  Bottom: the right hand never enters the camera frustum across the entire clip, leaving its channel unobserved at every frame.
  In both cases the generative process synthesizes a plausible but incorrect trajectory from the prior alone, illustrating the limit of our method when the upstream visual stream provides no observation to anchor a channel.}
  \label{fig:supp_failure_hot3d}
\end{figure*}

\begin{figure*}[!t]
  \centering
  \includegraphics[width=\linewidth]{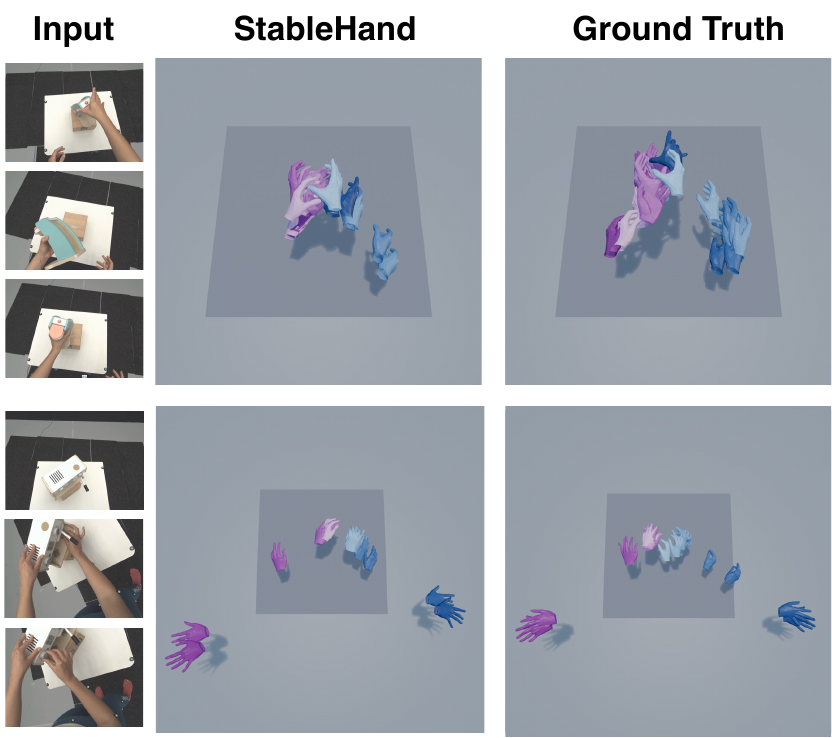}
  \caption{\textbf{Failure cases of StableHand on ARCTIC~\cite{fan2023arctic}.}
  Each row shows three input frames (left) together with our prediction and the Ground Truth, with color and temporal-shading conventions following Fig.~\ref{fig:qualitative}.
  Top: a bimanual contact phase in which the manipulated object simultaneously occludes both hands' fingers, leaving both finger channels with degraded observations across the contact span.
  Bottom: a bimanual manipulation phase in which the object covers both hands intermittently, leaving the two-hand spatial relationship without a reliable observation to anchor against.
  In both cases simultaneous degradation of both per-hand quality channels forces the generative process to recover from the prior, illustrating the structural limit of per-component quality conditioning when no hand provides a reliable observation across the contact phase.}
  \label{fig:supp_failure_arctic}
\end{figure*}

This section presents qualitative comparisons that complement Fig.~\ref{fig:qualitative} of the main paper.
Fig.~\ref{fig:supp_more_hot3d} and Fig.~\ref{fig:supp_more_arctic} report additional clips on HOT3D and ARCTIC respectively, following the same layout as the main qualitative figure.
Fig.~\ref{fig:supp_failure_hot3d} reports two representative failure modes of our method on HOT3D, in which one hand is unobserved by the upstream visual stream across most or all of the clip and the corresponding channel is synthesized entirely from the prior.
Fig.~\ref{fig:supp_failure_arctic} reports two analogous failure modes on ARCTIC, in which persistent bimanual hand-object occlusion simultaneously degrades both hands' observations during long contact phases.

\section[Qualitative Illustration of In-the-Wild Transfer]{\rev{Qualitative Illustration of In-the-Wild Transfer}}
\label{supp:in_the_wild}

\begin{figure}[!t]
  \centering
  \includegraphics[width=0.75\linewidth]{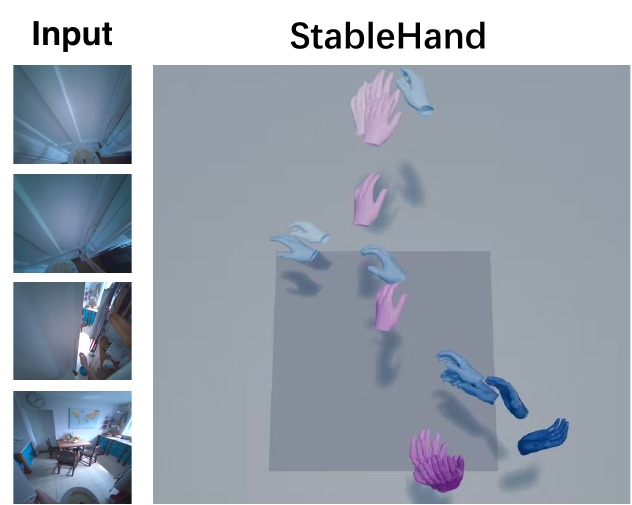}
  \caption{\textbf{\protect\del{Zero-shot in-the-wild inference}\protect\rev{Qualitative in-the-wild transfer} on HD-EPIC~\cite{perrett2025hd}.}
  Four input frames sampled from a single HD-EPIC clip (left) span dim corridors, kitchens, and dining rooms outside our training distribution.
  The right panel shows the recovered \protect\rev{world space} dual-hand mesh trajectory (left hand magenta, right hand blue, mesh shading dark$\to$light encoding temporal order).
  Neither the generative model (trained on HOT3D and ARCTIC) nor the quality network (pretrained on the eight-corpus mixture of Sec.~\ref{supp:datasets}) was exposed to HD-EPIC during training, yet the recovered trajectory remains spatially coherent.
}
  \label{fig:supp_in_the_wild}
\end{figure}

Beyond the in-distribution evaluations on HOT3D and ARCTIC, we run StableHand \del{zero-shot}\rev{without retraining} on HD-EPIC~\cite{perrett2025hd}, a recent in-the-wild egocentric corpus excluded from both the generative-model training splits and the eight-corpus quality-network pretraining mixture (Sec.~\ref{supp:datasets}).
Fig.~\ref{fig:supp_in_the_wild} reports a representative clip in which the egocentric camera traverses dim corridors and varied indoor scenes never seen at training time.
\rev{HD-EPIC has no 3D hand ground truth, and this example is a qualitative illustration rather than quantitative evidence.}
The recovered \rev{world space} dual-hand mesh trajectory remains spatially coherent\del{, suggesting that the per-component quality conditioning transfers to upstream observations whose error distribution differs from any single training corpus}\rev{, although the two hands interpenetrate in some frames}.

\end{document}